\documentclass[lettersize,journal]{IEEEtran}
\usepackage{amsmath,amsfonts}
\usepackage{algorithmic}
\usepackage{algorithm}
\usepackage{array}
\usepackage[caption=false,font=normalsize,labelfont=sf,textfont=sf]{subfig}
\usepackage{textcomp}
\usepackage{stfloats}
\usepackage{url}
\usepackage{verbatim}
\usepackage{graphicx}
\usepackage{cite}
\usepackage{color}
\usepackage{colortbl}
\usepackage{soul}
\usepackage{multirow}
\usepackage{multicol}
\usepackage{makecell}
\usepackage{booktabs}
\usepackage{hyperref}
\usepackage{booktabs}
\usepackage{amssymb}
\usepackage{bbding}
\usepackage{pifont}
\usepackage{wasysym}

\usepackage{amsfonts}
\usepackage{pifont}

\usepackage{amsmath}
 
\newcolumntype{P}[1]{>{\centering\arraybackslash}p{#1}}
\newcolumntype{C}[1]{>{\centering\arraybackslash}m{#1}}

\def \textHT [#1]{\color{red}$\mathbf{#1}$\color{black}}
\def \textLT [#1]{\color{blue}$\mathbf{#1}$\color{black}}

\newcommand{\YesMark}{\color{green}\textbf{\CheckmarkBold}\color{black}}
\newcommand{\NoMark}{\color{red}\textbf{\XSolidBrush}\color{black}}

\begin{document}

\title{General Place Recognition Survey: Towards the Real-world Autonomy Age}

\author{Peng Yin\textsuperscript{1,*},
% ~\IEEEmembership{Member,~IEEE},
        Shiqi Zhao\textsuperscript{2},
        Ivan Cisneros\textsuperscript{1}, 
        Abulikemu Abuduweili\textsuperscript{1}, 
        \\
        Guoquan Huang\textsuperscript{3}, 
        % ~\IEEEmembership{Senior Member,~IEEE},
        Micheal Milford\textsuperscript{4},
        % ~\IEEEmembership{Senior Member,~IEEE},
        Changliu Liu\textsuperscript{1},
        Howie Choset\textsuperscript{1},
        % ~\IEEEmembership{Fellow,~IEEE},
        and Sebastian Scherer\textsuperscript{1},
        % ~\IEEEmembership{Senior Member,~IEEE}
        % <-this % stops a space
\thanks{Peng Yin, Ivan Cisneros, Abulikemu Abuduweili, Changliu Liu, Howie Choset, and Sebastian Scherer are with the Robotics Institute, Carnegie Mellon University, Pittsburgh, PA 15213, USA. {(pyin2, icisnero, abulikea, cliu6, choset, basti)@andrew.cmu.edu}.}
\thanks{Shiqi Zhao is with the University of California San Diego, La Jolla, CA 92093, USA. {(s2zhao@eng.ucsd.edu)}.}
\thanks{Corresponding author: Peng Yin (pyin2@andrew.cmu.edu)}
}

\markboth{IEEE Transactions on Robotics (T-RO). Preprint Version. July, 2022}
{Yin \MakeLowercase{\textit{et al.}}: General Place Recognition Survey: Towards the Long-term Autonomy Age}

\maketitle

\begin{abstract}
Place recognition is the fundamental module that can assist Simultaneous Localization and Mapping (SLAM) in loop-closure detection and re-localization for long-term navigation. 
The place recognition community has made astonishing progress over the last $20$ years, and this has attracted widespread research interest and application in multiple fields such as computer vision and robotics. 
However, few methods have shown promising place recognition performance in complex real-world scenarios, where long-term and large-scale appearance changes usually result in failures.
Additionally, there is a lack of an integrated framework amongst the state-of-the-art methods that can handle all of the challenges in place recognition, which include appearance changes, viewpoint differences, robustness to unknown areas, and efficiency in real-world applications.
In this work, we survey the state-of-the-art methods that target long-term localization and discuss future directions and opportunities.

We start by investigating the formulation of place recognition in long-term autonomy and the major challenges in real-world environments.
We then review the recent works in place recognition for different sensor modalities and current strategies for dealing with various place recognition challenges.
Finally, we review the existing datasets for long-term localization and introduce our datasets and evaluation API for different approaches.
This paper can be a tutorial for researchers new to the place recognition community and those who care about long-term robotics autonomy.
We also provide our opinion on the frequently asked question in robotics: Do robots need accurate localization for long-term autonomy?
A summary of this work, as well as our datasets and evaluation API are publicly available to the robotics community at: 
\href{https://github.com/MetaSLAM/GPRS}{https://github.com/MetaSLAM/GPRS}.
\end{abstract}

\begin{IEEEkeywords}
Long-term Place Recognition, Multi-sensor modalities, Appearance-invariant, Viewpoint-invariant, Generalization Ability, Large-Scale Datasets
\end{IEEEkeywords}

\begingroup
\let\clearpage\relax
    \section{Introduction}
\label{sec:introduction}

    \begin{figure}[t]
        \begin{center}
            \includegraphics[width=\linewidth]{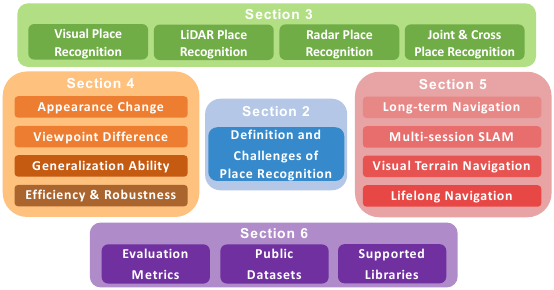}
        \end{center}
        \caption{\textbf{Structure of GPR Survey.}
        Place Recognition (PR) is the ability to recognize visited areas under different environmental conditions and various sensor modalities.
        In this survey, Section.~\ref{sec:definition_challenges} defines `place' and introduces the significant challenges within place recognition (PR);
        Section.~\ref{sec:sensor} investigates the PR approaches under different sensor modalities;
        Section.~\ref{sec:solution} and Section.~\ref{sec:application} provide the solutions for the current four major challenges and potential applications respectively; 
        finally, Section.~\ref{sec:data_eval} introduces the current datasets, metrics, and related supported libraries for place recognition research. 
        }
        \label{fig:idea}
    \end{figure}

    \IEEEPARstart{R}{eal}-world robotic systems have gained lots of attention in the past decades; from autonomous driving and last-mile delivery to search-and-rescue and warehouse logistics. Current robots are entering our daily lives and must deal with various complex scenarios.
    All of these mobile robotics systems require long-term localization to enable real-world autonomy.
    A critical aspect of long-term localization is guaranteeing the robust and accurate re-localization of previously visited areas under varying environmental conditions and viewpoint differences. This ability is referred to as place recognition (\textit{PR}) or loop closure detection (LCD).
    The area of \textit{PR} has been growing rapidly for the past ten years ($2012\sim 2022$): 
    1) there are more than $\mathbf{3400}$ published papers within this area, and 2) there have been several massive \textit{PR} competitions, such as the Google Landmark Recognition competition, the CVPR 2020 Long-term Visual Place Recognition competition, and our previous ICRA 2022 General Place Recognition competition for \href{https://www.aicrowd.com/challenges/icra2022-general-place-recognition-city-scale-ugv-localization}{City-scale UGV Localization} and \href{https://www.aicrowd.com/challenges/icra2022-general-place-recognition-visual-terrain-relative-navigation}{Visual Terrain Relative Navigation}.
    Despite the wealth of place recognition research, we see few methods that support robotic autonomy in long-term and large-scale real-world applications.

    \begin{figure*}[t]
            \begin{center}
                \includegraphics[width=\linewidth]{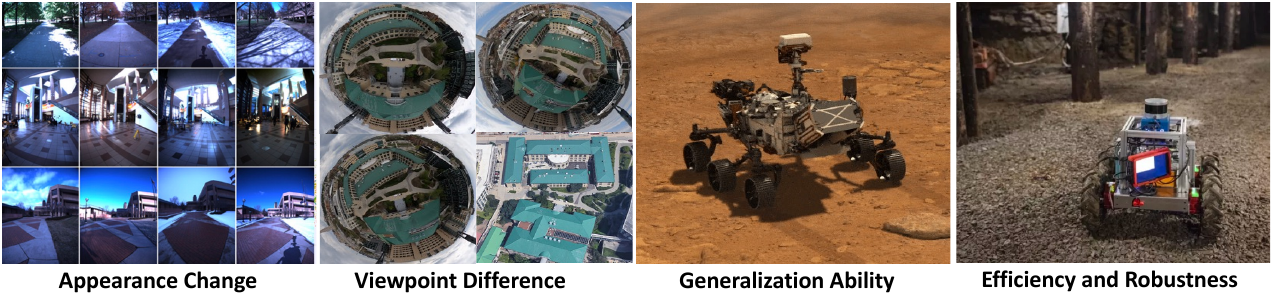}
            \end{center}
            \caption{\textbf{Challenges in Real-world Place Recognition.}
            In real-world navigation tasks, robots may encounter the following challenges: 
            1) changing visual appearances due to temporal variations (lighting, seasons), 
            2) diverse viewpoint differences for the same areas,      
            3) visiting new unknown areas,
            and 4) impacts to efficiency and robustness when applied to real-world scenarios.}
        \label{fig:issue}
    \end{figure*}
    
    \begin{table*} [t]
        \caption{Definitions (Types) of Place Recognition}
        \centering
        \begin{tabular}{|C{1.8cm}|C{7.5cm}|C{7.5cm}|}
            \hline
            & \textbf{Position-based}
            & \textbf{Overlap-based}
            \\
            \hline
            \textbf{Definition} &  
            The ability to recognize previously visited areas, irrespective of the viewpoint and conditional environmental differences.
            & 
            The ability to recognize previously visited areas based on the similarity of regions that overlap between the test and query images.
            \\
            \hline
            \textbf{Advantage} & 
            Suitable for place recognition, especially for large-scale SLAM tasks which need accurate loop closure detection. 
            & Suitable for content-/landmark- based place recognition and help develop place recognition methods without position restriction.
            \\
            \hline
            \textbf{Disadvantage} & Sensitive to position/viewpoint differences, and global environmental conditional differences. 
            &
            Hard to utilize for loop closure detection and will introduce localization noise under large-scale navigation tasks.
            \\
            \hline
        \end{tabular}
        \label{tab:defination}
    \end{table*}

    In previous works, Lowry \textit{et al.}~\cite{SURVEY:VPR} has provided a comprehensive analysis of visual place recognition (VPR) and defines the question of VPR as ``given an image of a place, can a human, animal, or robot decide whether or not this image is of a place it has already seen?";
    Zhang \textit{et al.}~\cite{Survey:VPR_Deep} give an updated survey of VPR with a focus on recent rapid developments in Deep Learning (DL);
    Rather than focusing on only the visual modality, Barros \textit{et al.}~\cite{Survey:PR_Deep} analyze place recognition using new sensor modalities (e.g., 3D LiDAR and RADAR) from the DL perspective.
    In~\cite{Survey:VPR_new}, Garg \textit{et al.} mentioned that ``the research on VPR has increasingly become more dissociated, there is no standard definition of a `place' ", and ``comparison of different methods is challenging as benchmark datasets and metrics vary substantially."
    In general, we also notice that it is hard to evaluate whether or not place recognition methods can be successfully applied to robotic systems for long-term autonomy.
    
    In light of the current dissociated status of PR, we aim to reduce the gap between theory and real-world application in this survey paper by analyzing the role of place recognition, describe its primary challenges, review the current solutions, and examine the potential research trends.
    Firstly, we re-formulate the problems and significant challenges for PR tasks in Section.~\ref{sec:definition_challenges}, where we argue the target for PR is to help improve the localization or data-association ability in long-term autonomy.
    Secondly, we analyze the current localization approaches across different sensor modalities (e.g., different visual inputs, LiDARs and RADARs) in Section~\ref{sec:sensor}, and provide a detailed comparison of their advantages and disadvantages.
    Thirdly, we investigate the details of existing solutions for PR tasks in Section.~\ref{sec:solution} and analyze their  properties, such as condition-invariance, viewpoint-invariance, generalization ability, and lifelong application.
    Then, we dive deep into the status of current PR applications in real-world autonomy in Section.~\ref{sec:application} and highlight the potential new trends and opportunities in large-scale and long-term navigation~\cite{intro:autonomous_car, intro:last_mile}, visual terrain relative navigation~\cite{VPR:SR_Season,grelsson2020gps},  multi-session SLAM~\cite{van2018collaborative,kimera}, and lifelong autonomy~\cite{mactavish2017visual,doan2020hm}.
    Though there are many place recognition datasets, one major problem is that it is either too complex (very complicated scenarios) or too simple (minimal differences) to evaluate PR performance with these datasets.
    To this end, we investigate the most recent public datasets that support real-world autonomy in Section.~\ref{sec:data_eval} and also provide the corresponding python library tools at \href{https://github.com/MetaSLAM/GPRS}{https://github.com/MetaSLAM/GPRS} to help evaluate localization performance of PR methods.
    Finally, in Section.~\ref{sec:conclusion} we discuss the status of current place recognition research from the perspective of real-world autonomy, which also serves as a reference for how to integrate PR with other relevant robotic and computer vision topics.
    In summary, the major contributions of this survey paper are as follows:
    \begin{itemize}
        \item This work provides a comprehensive discussion of the current status of place recognition from the following three perspectives:
        \textit{Uniform Problem Formulation}, \textit{Sensor Modalities}, and \textit{Challenges and Solutions}.
        We analyze both advantages and disadvantages of different solutions and modalities.
        %to other performance properties of PR.
        \item This work gives a deep dive into the \textit{Applications and Trends} in real-world autonomy, which includes large-scale and long-term mobile robot localization, visual terrain relative navigation, multi-agent localization, lifelong autonomy, etc.
        \item This work also provides a critical investigation of the \textit{Supporting Datasets and Evaluation Metrics} and highlights the current limitations for reasonable evaluation of real-world localization requirements. 
        We also provide our public datasets and open-source tools for fair comparison and analysis of future works.
    \end{itemize}
    \section{Definition of ``Place'' and Challenges}
\label{sec:definition_challenges}

In nature, even for tiny creatures (insects, birds, bats, mice, etc.), the ability of place recognition plays a important role in localization and navigation in a complex world.
Achieving robust localization ability also becomes one of the primary challenges for long-term real-world robotics autonomy.
The fundamental questions for place recognition in real-world autonomy are 1) ``what is a \textit{place}?'' and 2) ``what are the major challenges in place recognition tasks?''.

There primarily exist two definitions for place recognition: \textit{position-based} and \textit{overlap-based}.
In~\cite{SURVEY:VPR}, Lowry \textit{et al.} define the \textit{position-based} place recognition as the ability to recognize previously visited areas, irrespective to the viewpoint and environmental conditional differences.
The ``position-based" definition is inspired by the "place-cell" in the hippocampus, which was discovered by O'Keefe~\cite{place_cell} and helped him win the 2014's Nobel Prize.
In contrast, Garg \textit{et al.}~\cite{Survey:VPR_new} define the \textit{overlap-based} PR, which estimates retrievals' conditions using observation overlaps.
By this definition, the exact location under significantly different viewpoints will be treated as additional areas.
The \textit{overlap-based} definition is similar to the content-based image retrieval task~\cite{google_landmark} within the computer vision community, which aims at searching an extensive database for the most similar image to a given query image.
Due to this, though, \textit{overlap-based} PR cannot be directly used in a SLAM framework's loop closure detection for back-end position optimization~\cite{SLAM3D_survey}.

    \begin{figure*}[t]
        \begin{center}
            \includegraphics[width=0.95\linewidth]{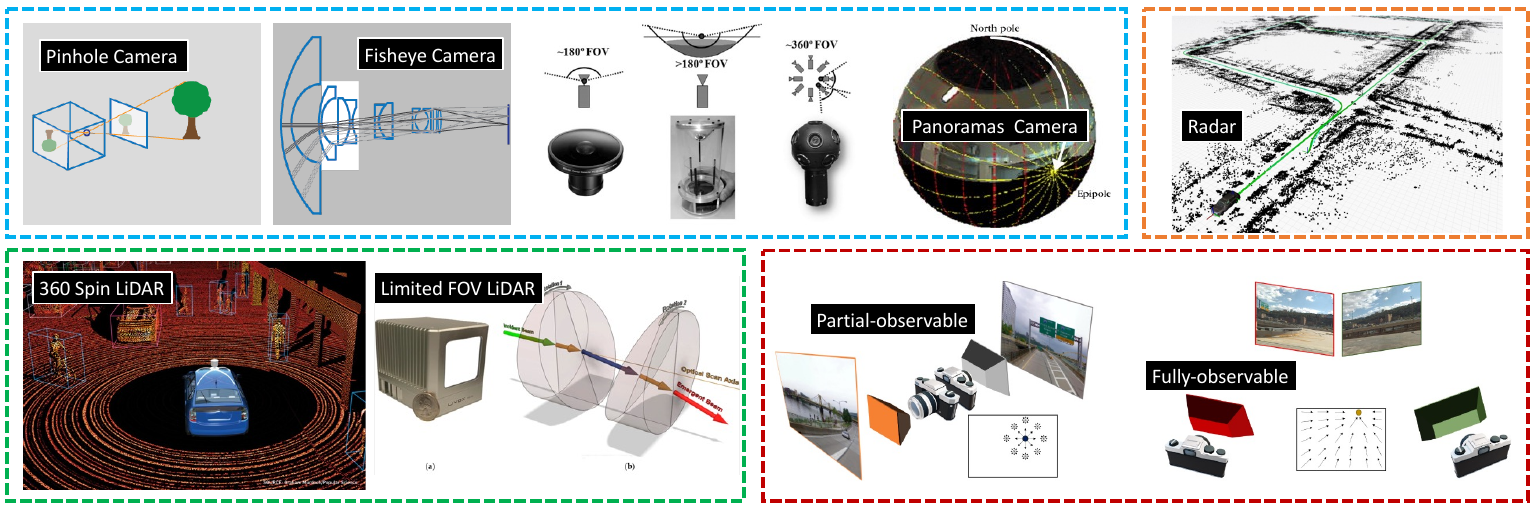}
        \end{center}
        \caption{\textbf{Diverse Sensor Modalities and Observation Properties}
        In the above figure, the blue box contains different camera setups, (a) pinhole camera, (b) dioptric camera (e.g. fisheye), (c) catadioptric camera, and (d) omnidirectional camera;
        the green box shows the two major LiDAR types: (a) the fully directional Spin LiDAR, (b) the limited field-of-view but non-repetitive LiDAR;
        The yellow box shows the modality of the Radar sensor for autonomous driving.
        Based on field-of-view, the sensors modalities can be categorized into two types, as shown in the red dashed box:
        (1) partial-observable sensor, such as pinhole/fisheye cameras, limited FoV LiDAR, and (2) fully-observable sensor, such as omnidirectional cameras and 360 Spin LiDAR and Radar devices.
        }
        \label{fig:sensor}
    \end{figure*}

In real-world robot autonomy, appearance changes can significantly affect place recognition performance.
As shown in Figure.~\ref{fig:issue}, we can summarize the major challenges of place recognition into four categories:
appearance change, viewpoint difference, generalization, and continual learning in real-world scenarios.
Compared to short-term navigation, long-term operation may contain appearance changes under different illumination conditions or structural changes (i.e., parking lot and construction sites), which will introduce further localization failures.
The viewpoint difference of the same area will decrease the place recognition accuracy in re-localization.
In large-scale navigation tasks, since the complex scenarios are unlimited, the generalization ability of place recognition methods for unseen environments is one of the primary burdens.
Beyond the generalization ability, the place recognition module is expected to have an online learning ability.
With the developments in space exploration, more robots have been sent to other plants for long-term investigation.
However, in such areas, human intervention is expensive, and the robots are expected to obtain continual learning ability for unseen areas.
Using the above definitions and challenges~\cite{SURVEY:VPR,Survey:PR_Deep}, we make a comparison between the two definitions in Table.~\ref{tab:defination}.
It indicates that both types of place recognition face challenges in the current navigation pipeline.

We argue that, though \textit{overlap-based} approaches are suitable for existing visual retrieval methods within the computer vision community, they are difficult to use in real-world applications.
For example, how do we determine the scale of overlaps? And how do we leverage the visual retrieval for loop closure detection in a SLAM framework?
It is also difficult to build the connection between \textit{overlap-based} place recognition and real-world localization and navigation requirements since we cannot estimate the relative positions between different images based only on overlaps.
In this case, the original \textit{position-based} place recognition is more suitable for real-world robot autonomy applications.
In this work, we will mainly focus on the related works that use a \textit{position-based} place recognition approach using different sensor types, solutions, and directions.
    \section{Sensor Modalities in Place Recognition}
\label{sec:sensor}

The first essential module for real-world robot localization is the appropriate sensor for the relative place recognition tasks.
Since different sensors have their own unique properties, place recognition performance depends highly on the adaptability of various sensors.
The critical considerations for sensor selection include 1) field of view, 2) geometric richness, and 3) robustness to environmental noise.
This section will look at place recognition methods based on different sensor setups.
And in Table.~\ref{tab:sensor}, we list the relative works for different sensor modalities and analyze their properties.

\begin{table*} [t]
\caption{Different Types of Camera Modalities}
\centering
\begin{tabular}{|C{3.5cm}|C{2cm}|C{11cm}|}
    \hline
    \textbf{Sensor}
    & \textbf{Related Work}
    & \textbf{Place Recognition Properties}
    \\
    \hline
    \textit{Partial-observable} Camera & \cite{Zaffar2020CoHoG,oertel2020augmenting,Khaliq2020RegionVLAD,ye2021visual,ozdemir2022echovpr,molloy2020intelligent,peng2021semantic,lu2021sta,waheed2021improving,khaliq2022multires,yuan2021softmp,piasco2021improving,mo2020fast,lee2021eventvlad} & These cameras are widely applied in most kinds of visual place recognition methods, but are sensitive to viewpoint differences, appearance changes, and dynamic objects.\\ 
    \hline
    \textit{Fully-observable} Camera  & \cite{yin2021i3dloc,yoon2014robust,lee2017fast,zioulis2021single,zhang2021panoramic,stone2016skyline,grelsson2020gps,liu2019lending} & 360-degree cameras provide rich geometries, and can provide viewpoint (orientation)-invariant place recognition, but are sensitive to appearance changes and dynamic objects.\\
    \hline
    LiDAR Device & \cite{Kim2018scancontext,guo2019local,zhu2020gosmatch,siva2020voxel,kim2020mulran,rozenberszki2020lol,komorowski2021minkloc3d,zywanowski2021minkloc3d,kim2021scan,yin2021pse,Yin2021spherevlad,vidanapathirana2021locus,yin2021fusionvlad,Xie2021RDC,li2022rinet,vid2022logg3d,lin2019fast,Wang2021Livox} & LiDAR provides the most accurate 3D environment measurements, and is invariant to appearance changes, but it usually has lower resolution and textureless inputs for confined/flat areas.\\
    \hline
    RADAR Device & \cite{gadd2020look,tang2020self,suaftescu2020kidnapped,gadd2021contrastive,cai2021autoplace,burnett2022radar,yin2021rall,hong2022radarslam} & Radar has 360-degree-view and not sensitive to both appearance and extreme weather conditions, but it has very low resolution inputs and noise cluster observations.\\
    \hline
    Joint Sensors & \cite{di2021visual,shan2021robust,yu2022mmdf,bernreiter2021spherical,cattaneo2020global,yin2021i3dloc,lai2021adafusion} & Joint sensors can cover the shortages of individual sensors, and the shared geometric property also make cross-modal place recognition methods possible.\\
    \hline
\end{tabular}
\label{tab:sensor}
\end{table*}

\subsection{Visual-based Place Recognition}
\label{sec:vpr}
Visual-based place recognition is the most investigated technique in the place recognition area.
As stated in the previous visual place recognition (VPR) surveys~\cite{SURVEY:VPR,Survey:VPR_new,Survey:VPR_Deep}, this area has been well studied using both traditional and learning-based approaches.
This section tries to avoid repeating the previous analysis of existing VPR approaches. Instead, we highlight the new VPR approaches based on different sensor modalities.
We argue that the same feature extraction approaches under different modalities may significantly differ in performance.
Based on the field-of-view, the camera modalities can be categorized into two types: 1) \textit{partial-observable} camera and 2) \textit{fully-observable} camera.

As we can see in Figure.~\ref{fig:sensor}.A, the \textit{partial-observable} camera mainly contains pin-hole, fish-eye, and stereo cameras.
\textit{partial-observable} camera systems are the most widely used device in visual place recognition tasks, and most \textit{overlap-based} place recognition approaches~\cite{VPR_Bench} and datasets~\cite{Survey:VPR_new} are conducted under this modality.
In~\cite{Zaffar2020CoHoG}, Zaffar~\textit{et al.} provide a non-learning method, CoHOG, which utilizes image entropy to extract regions-of-interest (ROI) and a regional-convolutional descriptor for appearance-/viewpoint-invariant visual place recognition.
Oertel~\textit{et al.}~\cite{oertel2020augmenting} provide an augmented VPR method for the pin-hole-based visual inputs, which utilizes both raw 2D visual inputs and 3D point-cloud from structure-from-motion to enhance the place recognition robustness.
Khaliq~\textit{et al.}~\cite{Khaliq2020RegionVLAD} provide a Region-based VLAD VPR method for significant viewpoint and appearance changes.
The critical contribution of this work is the lightweight CNN-based region features for fast and accurate global descriptor extraction.
Ye~\textit{et al.}~\cite{ye2021visual} propose a coarse-to-fine visual place recognition method, which extracts the global place descriptor from~\cite{sarlin2019coarse} for global matching, and parallelly with the local place descriptor for the local affine preserving matching (LAP). 
The proposed LAP method can provide effective and efficient feature matching for real-time geometrical verification of potential candidates.
In~\cite{ozdemir2022echovpr}, Ozdemir~\textit{et al.} introduce an Echo State Network (ESN)~\cite{jaeger2007optimization} enhanced method to model the spatial-temporal changes in VPR task, which can boost the VPR performance on different public datasets and shows the ability to capture temporal changes.
Against the appearance changing problem, Molloy~\textit{et al.}~\cite{molloy2020intelligent} provide the doubled Bayesian Selection Fusion method to determine the best potential matched candidates among the potential searching database under variant environmental conditions.
In~\cite{peng2021semantic}, Peng~\textit{et al.} introduced semantic reinforcement attention as the feature embedding layer to enhance the recognition performance with both semantic priors and data-driven fine-tuning, which can significantly improve the image representation.
To deal with the viewpoint difference in VPR, Lu~\textit{et al.}, ~\cite{lu2021sta} provides a Spatio-Temporal Alignment method, which contains an adaptive dynamic time warping (DTM) mechanism to align and measure the similarity of potential matches.
Waheed~\textit{et al.}~\cite{waheed2021improving} design a multi-process fusion method to fuse different VPR methods parallel for better performance.
Most recently, Lee~\textit{et al.}~\cite{lee2021eventvlad} utilized the event camera to provide reconstructed edge-based place recognition, and the unique representation of the event camera helps extract invariant structures under dynamic luminance changing environments.

Even though the above approaches using a \textit{partial-observable} camera can provide robustness against both viewpoint and appearance variants, the place recognition ability for the same area under different perspectives is still a significant challenge.
As we can see in Figure.~\ref{fig:sensor}.C, the limited field-of-view of a \textit{partial-observable} camera will cause it to observe significantly different objects from different traversal directions;
meanwhile, for a \textit{fully-observable} camera system, an observation with a 360-degree field-of-view will contain observations from all directions.
This property enables a \textit{fully-observable} camera system to have the innate advantage of viewpoint-invariant localization.
There are mainly two types of omnidirectional cameras, as stated in~\cite{scaramuzza2014omnidirectional}: catadioptric camera and panoramas camera.
Specifically, the catadioptric camera usually contains a fisheye camera (along with reflection mirrors in~\cite{yoon2014robust}) with upward directions, and the panoramas camera is combined with two fisheye cameras with inverse directions.
In~\cite{yoon2014robust}, Yoon~\textit{et al.} provides a catadioptric camera-based VPR method, which utilizes a spectral graph matching for flattened indoor environment navigation.
For aggressively maneuvering robot navigation on uneven terrain, Stone~\textit{et al.}~\cite{stone2016skyline} propose a Skyline-based localization system, which utilizes the spherical harmonics representation to provide viewpoint-invariant place recognition.
Zioulis~\textit{et al.}~\cite{zioulis2021single} provide a spherical panorama-based indoor Manhattan aligned layout estimation method, which can indirectly help the indoor localization task but has the basic cuboids assumption for indoor environments.
However, the field-of-view of a catadioptric configuration is limited, and the differences in the perspectives under significant pitch/roll changes will reduce the re-localization ability.
In~\cite{zhang2021panoramic}, the investigation by Zhang~\textit{et al.} shows that a 360-degree panoramic camera can provide more visual features in the visual SLAM system for odometry estimation and loop closure detection.
Based on the rich feature property of a 360-degree panoramic camera, Grelsson~\textit{et al.}\cite{grelsson2020gps} propose a position estimation method for unmanned surface vessels (USVs); the most exciting point of this work is that they can handle significant viewpoint differences in pitch/roll/yaw.
Their system combines a horizon line detection module to help approximate orientation estimation and a filter module to estimate the best matching localization in the Fourier domain.
Liu~\textit{et al.}~\cite{liu2019lending} provides a cross-view Geo-localization method, which contains a Siamese network to encode both 360 panorama images and the Google satellite images for geo-relative localization.
However, in the above applications, the localization robustness against viewpoint differences and the distortion problem caused by pitch/roll changes significantly affect the extracted visual features.
To deal with this problem, Yin~\textit{et al.}~\cite{yin2021i3dloc} provide an orientation-invariant place feature extraction method, which adds the orientation-equivalent property of spherical harmonics~\cite{Sphere:SO3_learning} to viewpoint-invariant place descriptors.

\begin{figure}[t]
	\centering
	\includegraphics[width=\linewidth]{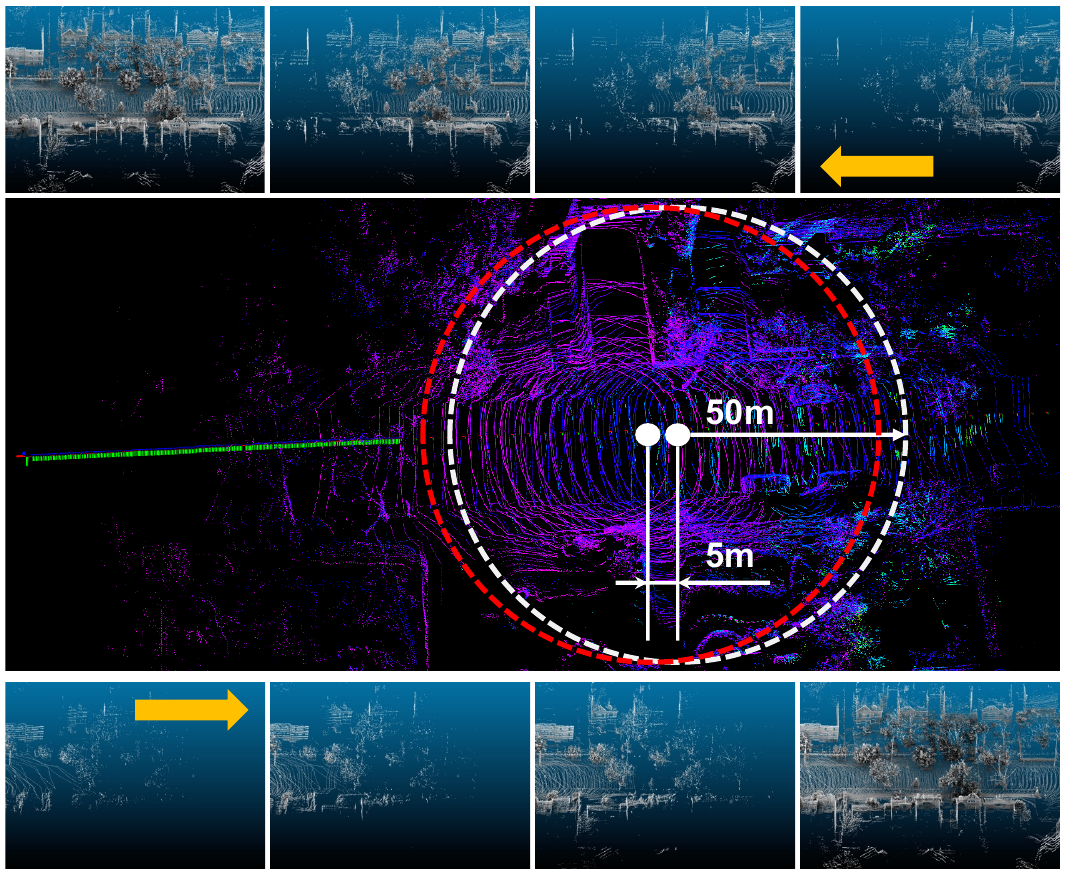}
	\caption{\textbf{Dense Sub-maps in LPR.}
        This figure shows the mapping results for the same area in the forward and backward directions.
	}
	\label{fig:dense_map}
\end{figure}

\subsection{LiDAR-based Place Recognition}
\label{sec:lpr}

In contrast to vision sensors, LiDAR-based devices can capture the 3D geometric structure of the surrounding environments via laser beams, which makes LiDAR-based place recognition (LPR) invariant to conditional differences, such as illumination, weather, and seasons.
As shown in the Figure.~\ref{fig:sensor}.C, there are two kind of LiDAR devices that are widely available, \textit{Spin} LiDAR (Velodyne VLP, 32E, 64E, etc~\cite{moosmann2011velodyne}) and \textit{Non-repetitive} LiDAR (Livox LiDAR, Mid-40, AVIA, etc~\cite{lin2020loam}).
The \textit{Spin} LiDAR devices usually have $100\sim120m$ measurement range and 360-degree field-of-view but are also expensive and have low-resolution sensor inputs;
in contrast, the \textit{Non-repetitive} LiDAR devices are relatively cheap and can provide high-resolution inputs, but their field of view is fairly limited.
This subsection will introduce the related works in LPR.

Based on the development of Deep 3D feature extraction, Uy~\textit{et al.}~\cite{uy2018pointnetvlad} provide 3D LPR method by combining the PointNetVLAD~\cite{qi2017pointnet} feature extraction approach and NetVLAD~\cite{NetVLAD} feature cluster method, which transform the successful visual place recognition technique into LiDAR domain.
However, different from visual inputs, the low-resolution raw LiDAR inputs make it hard to encode local features into the distinguishable global descriptor.
Guo~\textit{et al.}~\cite{guo2019local} design an enhanced place descriptor by combining both LiDAR point cloud and its calibrated intensity returns, showing significant performance improvements over pure geometric-based methods.
Zhang~\textit{et al.}~\cite{zhang2019pcan} propose an attention~\cite{vaswani2017attention} enhanced place feature extractor, which can predict the significance of local points based on their context.
To enhance the connection of local 3D structures in the large-scale point cloud, Liu~\textit{et al.}~\cite{liu2019lpd} introduces a graph-based neighborhood aggregation module to reveal the spatial distribution within the global descriptor.
Zhu~\textit{et al.}~\cite{zhu2020gosmatch} provide a coarse-to-fine strategy to perform semantic-level rough alignment and accurate pose refinements with 3D geometric registration. 
To deal with the sparsity LiDAR scan in long-term place recognition, Siva~\textit{et al.}~\cite{siva2020voxel} developed a voxel-based representation learning method.
Yin~\textit{et al.}~\cite{yin2021pse} utilize a parallel semantic embedding module to encode different kinds of objects (ground, building, small structures, etc.) and a divergency learning module to learn more robust 3D place descriptors under long-term variants.
To deal with the sparsity of a single scan and provide a consistent place descriptor, Yin~\textit{et al.}~\cite{yin2021fusionvlad} accumulate LiDAR scans into dense sub-maps for LPR feature extraction.
As shown in Figure.~\ref{fig:dense_map}, the generated dense sup-map can improve the LiDAR map similarity under inverse directions, and similar approaches have also been applied in~\cite{uy2018pointnetvlad,liu2019lpd,komorowski2021minkloc3d}.
Vidanapathirana~\textit{et al.}~\cite{vidanapathirana2021locus} utilize a second-order pooling to aggregate the topological and temporal semantic information to the place descriptor, which shows higher robustness to viewpoint changes and occlusions.

However, the PointNet~\cite{qi2017pointnet} structure is not suitable to capture local geometric structures, especially under orientation differences, which will cause a performance drop under viewpoint differences, as mentioned in~\cite{liu2019lpd,vidanapathirana2021locus}.
In~\cite{Kim2018scancontext}, Kim~\textit{et al.} provides a rotation-invariant 3D place descriptor through the structural appearance of polar context projection; however, this method is robust to orientation difference but very sensitive to local translation difference due to polar representation.
Li~\textit{et al.}~\cite{li2022rinet} propose a rotation invariant place descriptor via a series rotation-invariant network structures (convolution, attention, down-sampling).
In~\cite{Yin2021spherevlad}, Yin~\textit{et al.} provide a spherical harmonica-based 3D map encoding method, the rotation-equivalent property of spherical harmonica~\cite{Sphere:SO3_learning} enhance the extracted place descriptors are invariant to local viewpoint differences.
As an extension of~\cite{Kim2018scancontext}, Kim~\textit{et al.}~\cite{kim2021scan} provides a structural appearance-based LPR method, the new method can extract both translation- and rotation- invariant LiDAR place descriptor simultaneously.
In~\cite{yin2021fusionvlad}, Yin~\textit{et al.} provides a fusion-based multiview place feature encoding method, which can combine the translation-invariant property of the top-down view and the rotation-invariant property of the spherical view and give an even robust 3D descriptor.
Komorowski~\textit{et al.}~\cite{komorowski2021minkloc3d} introduce a voxelization representation for the 3D LPR task and utilize sparse 3D convolution for place descriptor extraction, which has shown significant improvements to PointNet-based methods.
As an extension of~\cite{komorowski2021minkloc3d}, Zywanowski~\textit{et al.}~\cite{zywanowski2021minkloc3d} utilize spherical coordinates of 3D points and LiDAR intensity measurements to further improve the recognition robustness with even a single sparse LiDAR scan.
Vidanapathirana~\textit{et al.}~\cite{vid2022logg3d} also utilize the sparse convolution in LPR descriptor extraction and provides a local consistency loss to guide the network to learn consistent local features across revisits, which also significantly improves the re-localization accuracy compared to other sparse convolution-/PointNet- based LPR approaches.
For the \textit{Non-repetitive} LiDAR device, Lin~\textit{et al.}~\cite{lin2019fast} develop a fast loop detection approach  where they calculate the cross-correlation of 2D histograms to estimate keyframes' similarities.
Wang~\textit{et al.}~\cite{Wang2021Livox} utilize the ScanContext~\cite{Kim2018scancontext} in the \textit{Non-repetitive} scans for place recognition.

\subsection{Radar-based Place Recognition}
\label{sec:rpr}

Different from camera and LiDAR sensors, which are operated under near-visible electromagnetic spectra (higher frequency, from THz to PHz), Frequency-Modulated Continuous Wave (FMCW) scanning Radar devices are usually performed under much lower frequency (GHz)~\cite{hong2022radarslam}.
The property of the Radar device makes it not sensitive to illumination differences, appearance changes, and extreme weather conditions (i.e., fog, mist, rain, snow, etc.).
And currently, available Radar devices can also provide 360-degree field-of-view with up to hundreds of measurement abilities.
The above properties make Radar-based place recognition (RPR) suitable for challenging outdoor localization under extreme weather conditions~\cite{burnett2022radar}. 
In~\cite{hong2022radarslam}, Hong~\textit{et al.} provide a complete radar SLAM system, which includes the motion estimation for reliable feature tracking, a motion distortion module to refine the pose estimation, and an M2DP~\cite{he2016m2dp} based place recognition module to provide efficient loop closure detection for large-scale SLAM tasks.

However, Radar devices have a typical limitation compared to a camera and LiDAR devices, low spatial resolution and noise cluster observations, making Radar-based place recognition still very challenging.
In~\cite{gadd2020look}, Gadd~\textit{et al.} provides a sequence matching-based RPR method against the noise clusters within single scans, which can achieve a $30\%$ performance boost on $26$km urban environments.
However, in the real-world RPR task, the relative datasets are limited compared to camera-/LiDAR- based ones.
Tang~\textit{et al.}~\cite{tang2020self} provide a self-supervised fusion method for RPR task, which is cheap to train without accurate ground truth. 
Specifically, they utilize public satellite images to simulate the Radar projections for metric localization learning.
In~\cite{cai2021autoplace}, Cai~\textit{et al.} develop the dynamic points removal module to reduce the vibrant noise and the deep spatial-temporal feature embedding module to improve the extracted Radar points quality.
Like the 360-degree LiDAR device, rotation differences may also encounter viewpoint differences in the RPR large-scale localization.
Suaftescu~\textit{et al.}~\cite{suaftescu2020kidnapped} provide a rotation-invariant radar-based place recognition method, which utilizes the cylindrical convolutions, anti-aliasing blurring, and azimuth-wise max-pooling to improve the recognition accuracy for polar radar scans in large-scale outdoor localization task.
Gadd~\textit{et al.}~\cite{gadd2021contrastive} provide a data augmentation approach in the place descriptor training procedure, which can further improve the RPR performance.

\subsection{Joint and Cross Place Recognition}
\label{sec:jcpr}

As we summarized in Table.~\ref{tab:sensor}:
vision sensors can provide rich texture and semantic information but are sensitive to illuminations and environmental condition differences;
LiDAR devices are robust to the above appearance changes but have low-resolution data outputs and are sensitive to occlusions;
Radar devices give even longer detections than LiDAR and are robust to extreme weather conditions but contain lots of measurement noise.
Each sensor modality has unique adaptive properties (i.e., long-range in visual sensor, illumination-invariant in LiDAR sensor, and dust-invariant in Radar sensor), distinguishes abilities under different environmental conditions, and shares commonalities.
Based on the properties mentioned above, there mainly exits two kinds of multi-modality place recognition approaches:
1) for the purpose of robust general place recognition, the combination of multi sensor modalities provides enhanced \textit{Joint} place recognition methods~\cite{shan2021robust,yu2022mmdf,bernreiter2021spherical}.
2) based on the commonality of different modalities on the same geometric structures, the \textit{Cross} place recognition methods provide efficient and low-cost re-localization over existing maps~\cite{cattaneo2020global,di2021visual,yin2021rall,yin2021i3dloc}.

For the \textit{Joint} place recognition, Shan~\textit{et al.}~\cite{shan2021robust} developed an ORB descriptor based on both visual inputs and LiDAR intensity readings, which is invariant to rotation differences due to the panoramic perspective and RANSAC-based outlier rejecting.
In~\cite{yu2022mmdf}, to improve the navigation robustness under cross-scene (ground, water, surface) environments, Yu~\textit{et al.} provide a multi-modal deep feature fusion method for both LiDAR and image inputs. 
Specifically, they also utilize the semantic feature to enhance the point cloud feature. 
The final place descriptor is aggregated from a fusion layer, which shows higher robustness than the original LiDAR and image features. 
Similar to~\cite{shan2021robust}, Bernreiter~\textit{et al.}~\cite{bernreiter2021spherical} applies the spherical convolution neural network for the joint LiDAR and camera inputs to extract global place descriptor, which shows better recognition accuracy and robustness to viewpoint differences.
However, the place recognition performance of different sensor modalities may also depend on the relative environmental conditions, i.e., the visual sensor in the dark environment and the LiDAR sensor under flattened areas.
To deal with this problem, Lai~\textit{et al.}~\cite{lai2021adafusion} provides an end-to-end adaptive fusion approach, which can be based on the individual sensor performance under different conditions and adjust their feature weightings in the global place descriptor fusion.

For the \textit{Cross} place recognition, Cattaneo~\textit{et al.}~\cite{cattaneo2020global} provides a cross-domain visual localization within existing LiDAR maps, which trains a 2D VPR network and a 3D LPR network simultaneously and provides robust cross-domain localization under changing weather and lighting conditions.
In~\cite{di2021visual}, Di~\textit{et al.} utilize the stable intensity measurements from the LiDAR device, and provide a deep evaluation of the VPR method against existing LiDAR intensity data; 
the relative results on different public datasets suggest that visual-to-intensity-based place recognition can deliver reliable performance in large-scale visual navigation. 
Yin~\textit{et al.}~\cite{yin2021i3dloc} provide a condition- and viewpoint- invariant place recognition method based on a conditional domain transfer module and an orientation equivalent extraction module.
In~\cite{yin2021rall}, Yin~\textit{et al.} provide Radar localization method against the LiDAR maps, where both Radar and LiDAR maps are fed into the standard embedding networks to learn cross-matching features.
    \begin{figure*}[t]
    \begin{center}
        \includegraphics[width=\linewidth]{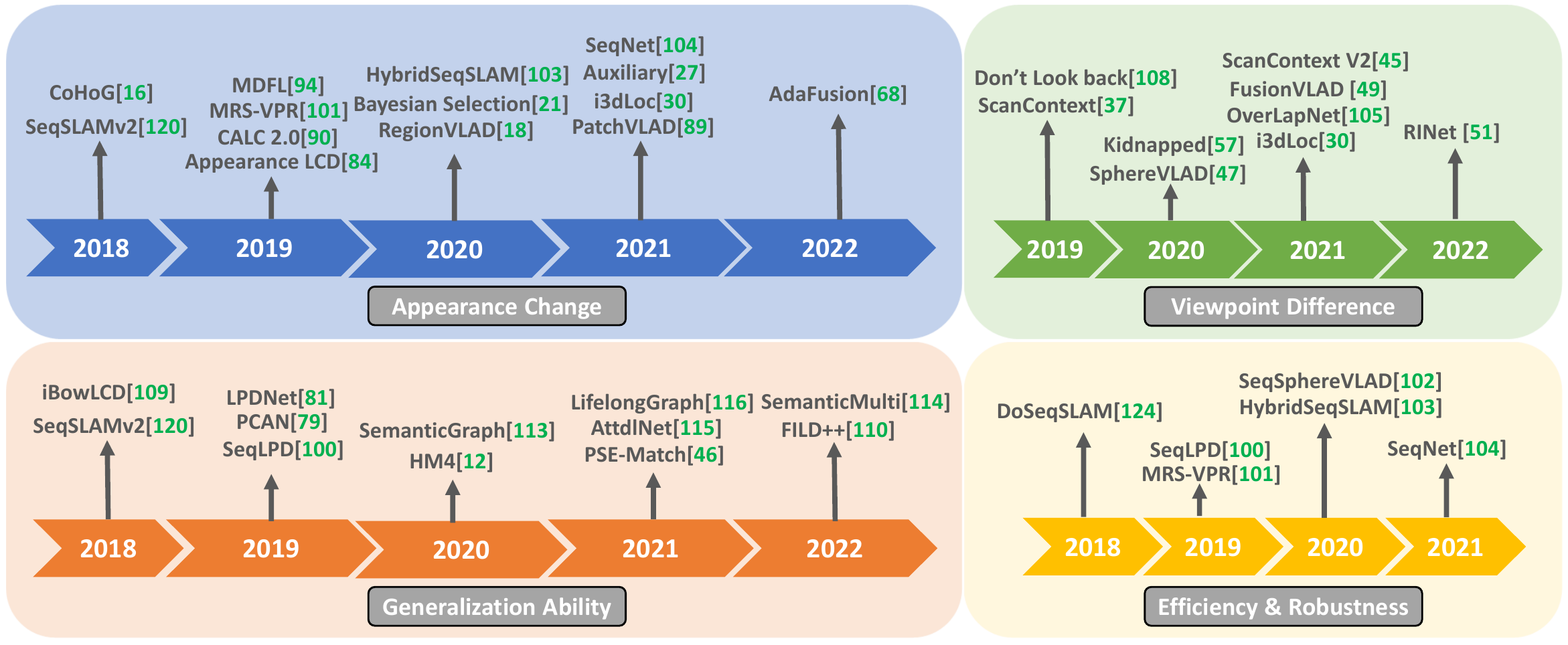}
    \end{center}
    \caption{\textbf{Overview of GPR works on Major Challenges.}
    This figure illustrates the trends of place recognition methods to deal with the significant challenges in Appearance Changing, Viewpoint Differences, Generalization Ability, and Efficiency \& Robustness from 2018 to 2022.
    }
    \label{fig:trends}
\end{figure*}

\section{Key Strategies in Place Recognition}
\label{sec:solution}
As we stated in Section.~\ref{sec:definition_challenges}, the primary challenges for place recognition can be categorized into four types: 
1) Appearance change, 2) Viewpoint difference, 3) Generalization ability, and 4) Lifelong performance.
We will investigate them separately.

\subsection{Appearance Change}
\label{sec:appearance_changes}
As one of the main challenges in place recognition, appearance changing can introduce recognition failures for the same areas and perceptual aliasing where different places generate similar visual observations.
In long-term place recognition, there are two types of appearance changes:
\begin{itemize}
    \item \textit{Conditional changes}, contains the appearance changes caused by environmental conditions, such as illuminations, weather, and seasons. 
    This type of change will mainly affect the visual observations over time.
    \item \textit{Structural changes}, contains the dynamic objects, geometric transformations, and landform changes over short-term or long-term navigation.
    Both vision and LiDAR sensors can be affected, but the Radar device can survive because of its low-frequency observation property~\cite{hong2022radarslam}.
\end{itemize}
As pointed in~\cite{Survey:VPR_Deep}, there mainly exist two kinds of approaches to deal with the above appearance changes, 1) \textit{place modeling}, which utilizes the conditional-invariant features for stable place recognition, and 2) \textit{belief generation}, which estimates the place similarity based on the sequence of observations.

With the rapid development in computer vision~\cite{feature:orb,tsintotas2019appearance,Zaffar2020CoHoG} deep learning~\cite{Feature:VGG,he2016identity}, and adversarial learning~\cite{goodfellow2014generative, zhu2017unpaired}, the \textit{place modeling} based PR methods~\cite{Khaliq2020RegionVLAD,piasco2021improving,hausler2021patch,yin2021i3dloc} have gained lots of attention.
In~\cite{tsintotas2019appearance}, Tsintotas~\textit{et al.} provides a training procedure to capture repeated scale-restrictive features and a voting mechanism to detect the potential loop closures within the distributed database.
Merrill~\textit{et al.}~\cite{merrill2019calc2} combines different visual modes information to extract rich place recognition features for robust re-localization.
Inspired by the success of region-based visual object detection, Khaliq~\textit{et al.}~\cite{Khaliq2020RegionVLAD} provides a Region-based VLAD feature aggregation module, which utilizes the pretrained AlexNet~\cite{VPR:ALEXNET} for stable local region feature extraction.
Hausler~\textit{et al.}~\cite{hausler2021patch} provides a multi-scale patch feature fusion mechanism to combine both local and global descriptions to boost the place recognition accuracy.
Hausler's method can aggregate local features over the feature-space grid, which is robust to environmental condition differences.
However, the performance of the above methods is highly dependent on the pretrained model on existing datasets, and the adaptation ability is limited.
Zaffar~\textit{et al.}~\cite{Zaffar2020CoHoG} provide a training-free VPR approach, which combines the traditional HOG~\cite{VPR:HoG} descriptor and the regional feature extraction and convolution matching mechanism to achieve competitive recognition performance without training.
Piasco~\textit{et al.}~\cite{piasco2021improving} introduce the auxiliary modality for the VPR task, which can infer depth prediction from the visual inputs and combine with the visual CNN features to extract robust global place descriptor.
However, this work assumes there are paired image-depth data. 
Thus, the localization performance and generalization ability are highly reliant on the dataset and the depth of image quality.
Based on the feature extraction ability of CapsuleNet~\cite{sabour2017dynamic}, Yin~\textit{et al.}~\cite{yin2019multi} provide a multi-domain feature learning method, which can capture different attribute descriptions from the diverse observations and separate the condition-invariant features for recognition.
As an extension of~\cite{yin2019multi}, Yin~\textit{et al.}~\cite{yin2021i3dloc} further provides a cross-domain transfer module, different from the previously fixed domain-transfer method~\cite{zhu2017unpaired}, this method introduces a conditional adversarial feature learning network to extract conditional-invariant visual features under multiple environmental conditions.
However, the complex variant environmental conditions make the place recognition still challenging even under the assistance of \textit{place modeling}.
Molloy~\textit{et al.}~\cite{molloy2020intelligent} combine the NetVLAD-based place feature extraction with a Bayesian selective fusion approach, which can find target matches under relative domains.
But this approach requires complete reference images under all domains. 

\begin{figure}[t]
    \centering
    \includegraphics[width=\linewidth]{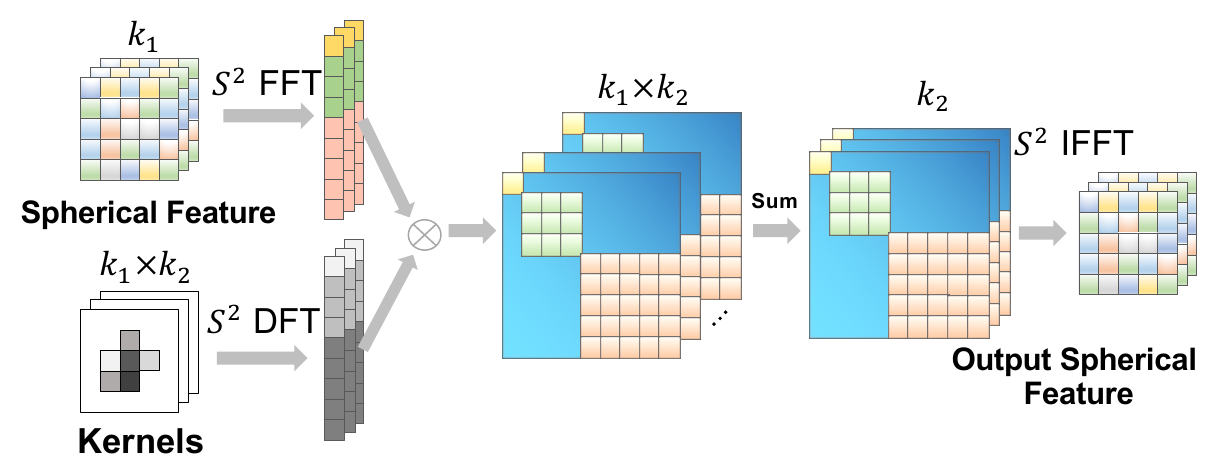}
    \caption{\textbf{Spherical Harmonic-based Convolution.}
    Given a spherical projection feature and a kernel signal, we can transform them into a harmonic domain~\cite{Yin2021spherevlad} using Fourier transform (Fast or Discrete) for convolution operations and get output features via inverse Fast Fourier transform (IFFT).}
    \label{fig:sphere_cnn}
\end{figure}

For \textit{belief generation} based PR methods, SeqSLAM~\cite{Milford2012SeqSLAM} introduced by Milford~\textit{et al.} provide the well-known sequence matching method. 
In contrast to the traditional single frame-based place recognition methods, such as FAB-MAP~\cite{cummins2008fab} and Bag-of-Words~\cite{VPR:DBOW2}, SeqSLAM finds the best matches by aligning a pair of reference and query sequence and can accurately capture the continuous geometric similarities under \textit{Conditional changes} even with most traditional visual features~\cite{VPR:HoG,feature:orb}.
Since $2012$, there exist a wide body of literature~\cite{stone2016skyline,Siam2017FastSeqSLAM,bampis2018fast,Liu2019SeqLPD,Yin2019MRS_VPR,Yin2020SeqSphereVLAD,Chancan2020HybridSeqSLAM,Garg2021SeqNet} benefiting from the SeqSLAM mechanism to provide more robust place recognition.
In~\cite{stone2016skyline}, to stable place recognition ability on a rough terrain under changing conditions, Stone~\textit{et al.} combines the SeqSLAM matching mechanism with the segments of skylines, which shows improved condition-invariant property during bumpy trails.
To improve the searching efficiency of SeqSLAM, Siam~\textit{et al.}~\cite{Siam2017FastSeqSLAM} provides a fast version of SeqSLAM, which utilize the approximate nearest neighbor (ANN) to greedily search the sequence instead of traditional burst-force searching and can significantly reduce the searching time without degrading the localization accuracy. 
However, this method initially needs good initial estimation, thus not suitable for large-scale global re-localization.
In~\cite{Yin2019MRS_VPR,Yin2020SeqSphereVLAD}, Yin introduces a multi-resolution sampling-based global place recognition mechanism for VPR~\cite{Yin2019MRS_VPR} and LPR~\cite{Yin2020SeqSphereVLAD}.
Yin's method combines the coarse-to-fine re-sampling with particle-filter to achieve hierarchically global re-localization; this approach can balance the matching accuracy and efficiency and help provide near real-time global localization ability in long-term navigation.
Instead of using burst-force searching over the difference matrix in SeqSLAM, Bampis~\textit{et al.}~\cite{bampis2018fast} provides a sequence bag-of-words with a novel temporal consistency filter, which can benefit from sequence matching manner but also maintain the real-time performance on a tablet device.
In~\cite{Chancan2020HybridSeqSLAM}, Chancan~\textit{et al.} combines a compact and sparse neural network, FlyNet, and a continuous attractor neural network (CANN) to capture the sequence of observations, which can provide a better performance than SeqSLAM.
Similarly, to~\cite{Chancan2020HybridSeqSLAM}, Garg~\textit{et al.}~\cite{Garg2021SeqNet} also provides sequence global descriptor by using a temporal convolutional network, which can construct a hybrid place recognition system by combining both global matching and the local sequence matching.

\begin{figure}[t]
    \centering
    \includegraphics[width=\linewidth]{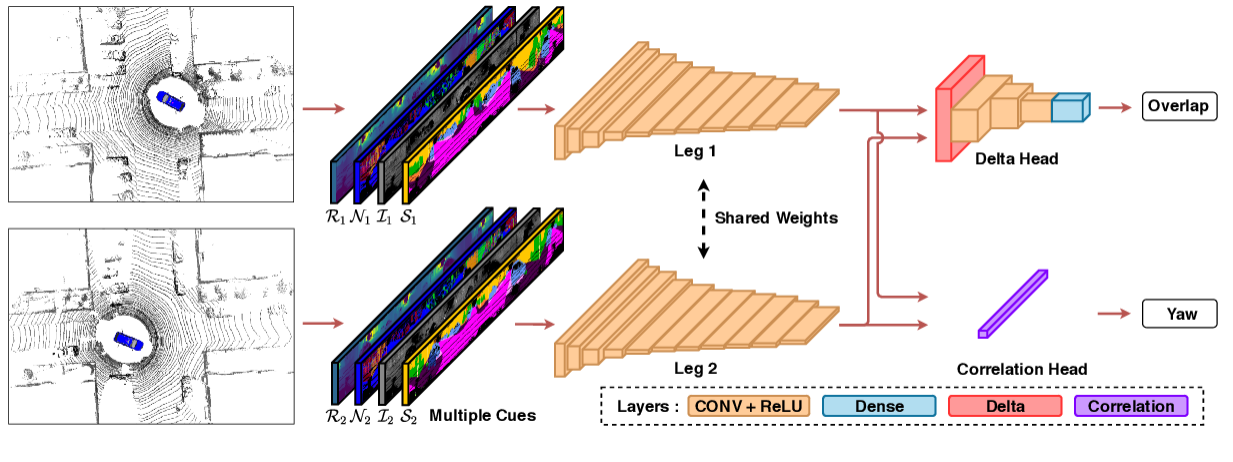}
    \caption{\textbf{Overlap-based Matching Estimation.}
    Given two-column views, the pairwise-based method~\cite{chen2021overlapnet} utilizes a shared weights convolution layer to extract local features and applies the delta operation to estimate the relative yaw angle and overlaps between two observations.}
    \label{fig:Overlap_est}
\end{figure}

\subsection{Viewpoint Difference}
\label{sec:view_diff}

As stated in Section.~\ref{sec:appearance_changes} viewpoint differences can affect the recognition ability for all kinds of sensor modalities as one of the significant challenges in place recognition. 
In long-term navigation, the observation in the same place may have variant patterns under different perspectives.

To improve the robustness of viewpoint difference, traditional place recognition methods, such as Bag-of-words~\cite{VPR:DBOW2}, Fisher Vector~\cite{sanchez2013image}, and Vector of Locally Aggregated Descriptors(VLAD)~\cite{VLAD}, have been investigated for more than ten years.
However, the above methods can not deal with arbitrary viewpoint differences or inverse directions. The distribution changes caused by viewpoints will be treated as different places in \textit{Position-based} place recognition.
Garg~\textit{et al.}~\cite{garg2018don} provide semantics-aware neural networks to recognize the same place under inverse directions.
However, this method can only deal with inverse-direction recognition, while arbitrary real-world differences are more common in indoor/outdoor navigation.
As analyzed in Section.~\ref{sec:sensor}, the panorama camera, LiDAR, and Radar can provide fully-observable sensor inputs, which are highly suitable for viewpoint invariant place recognition.
As we mentioned in Section.~\ref{sec:lpr}, Kim~\textit{et al.}~\cite{Kim2018scancontext,kim2021scan} provide a rotation-invariant descriptor through polar context projection; and Li~\textit{et al.}~\cite{li2022rinet} also propose a rotation-invariant place descriptor via the combination of rotation-invariant network structures.
In~\cite{Yin2020SeqSphereVLAD,yin2021i3dloc}, Yin~\textit{et al.} maps the LiDAR or 360-degree visual inputs to the spherical projections, and utilize the orientation-equivalent property of spherical harmonics to provide orientation-invariant place recognition methods as shown in Figure.~\ref{fig:sphere_cnn}.
Chen~\textit{et al.}~\cite{chen2021overlapnet} develop a deep neural network to estimate the overlap similarly based on the overlap estimation from two LiDAR scans and parallel estimate the relative yaw angles via the delta-operation as depicted in Figure.~\ref{fig:Overlap_est}.
However, the overlap-based methods need to calculate each potential data pair; this procedure can make training and evaluation procedures time-consuming compared to the global descriptor-based spherical harmonics methods.
Based on the dense local maps, Yin~\textit{et al.}~\cite{yin2021fusionvlad} provide a multi-perspective fusion-based LPR, which benefits from the translation-invariant property of top-down inputs and the orientation-invariant property from a spherical view.
In~\cite{suaftescu2020kidnapped}, Suaftescu~\textit{et al.} provides an orientation-invariant RPR method, which utilizes cylindrical convolutions along with anti-aliasing blurring, and azimuth-wise max-pooling to improve the localization accuracy for the noise Radar inputs.

\begin{figure*}[t]
    \begin{center}
        \includegraphics[width=\linewidth]{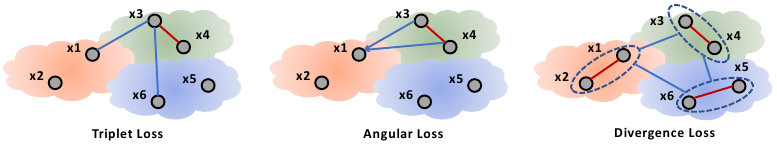}
    \end{center}
    \caption{\textbf{Loss Metric in Place Recognition.}
    The triplet loss (left) considers increasing features' Euclidean distances of different clusters and decreasing the distances within the same clusters.
    The angular loss (middle) considers using angular (i.e., cosine) distance instead of Euclidean distance, which is invariant to scale transform differences.
    The divergence loss (right) also considers the cluster-wise distance instead of only individual members' distances.
    }
    \label{fig:loss_metric}
\end{figure*}

\subsection{Generalization Ability}
\label{sec:generalization}
Generalization ability indicates the place recognition of unseen environments.
The real-world scenarios are infinite, given the same place can be presented under different environmental conditions, as stated in Section.~\ref{sec:appearance_changes}, there is no boundary for the above combination of place datasets, which is impossible to collect at once.

% hand-craft features
In traditional methods, FAB-MAP~\cite{cummins2008fab} build a Bag-of-visual-words (BoW) architecture to achieve large-scale visual re-localization.
iBoW-LCD~\cite{ibow_lcd} uses an incremental BoW scheme based on binary descriptors to retrieve matched images more efficiently.
An~\textit{et al.} introduces FILD++~\cite{bow_fast}, an incremental loop closure detection approach via constructing a hierarchical small‐world graph.
However, the generalization ability of the above non-learning-based methods is highly restricted, and robust performance usually comes with specific parameter fine-tuning.
% Deep learning features
With the development of deep learning-based feature extraction~\cite{Feature:VGG,resnet} and attention mechanism~\cite{vaswani2017attention}, learning-based place feature extraction methods have gained more attention in recent years.
Khaliq~\textit{et al.}~\cite{Khaliq2020RegionVLAD} combine the region of CNN features and a differentiable NetVLAD~\cite{NetVLAD} layer to enable the generalization ability of the VPR task.
Based on the attention mechanism, Zhang~\textit{et al.}~\cite{zhang2019pcan} proposes a Point Contextual Attention Network (PCAN) to enforce the differential networks by paying more attention to the task-relevant features, which can further boost the robustness of LPR methods.
In~\cite{lai2021adafusion}, Lai combines the advantages of different sensor modalities and proposes the attention-enhanced multi-modal fusion-based PR method, which can balance the feature weights based on the environmental conditions to boost the localization performance.
% Ma~\textit{et al.}~\cite{ma2022overlaptransformer} extend the original OverlapNet~\cite{chen2021overlapnet} with the attention module, which also boost the LPR performance in long-term place recognition.
In the most recent works, Zhao~\textit{et al.}~\cite{zhao2022spherevlad2} extend the original SphereVLAD~\cite{Yin2021spherevlad} with an attention mechanism to increase the signal-noise ratio of extracted place features, which can improve the global feature distinguishable ability in large-scale place recognition.

% Semantic enhanced method
On the other hand, the semantic segments in the 3D environments are usually more robust to the environmental changes.
Based on this property, Kong~\textit{et al.}~\cite{kong2020semantic} provide semantic graph-based place recognition methods and utilize a graph-matching to improve localization performance under occlusion and viewpoint differences.
Similarly, Yin~\textit{et al.}~\cite{yin2021pse} propose a parallel semantic feature encoding module, which extracts the different types of semantics (tree, building, etc.) and utilizes a divergence place learning network to improve the robustness of place recognition.
Most recently, Paolicelli~\textit{et al.}~\cite{paolicelli2022learning} combined the visual appearance and semantic context through a multi-scale attention module for robust feature embedding.
% New loss metrics based method

    \begin{figure*}[t]
            \begin{center}
              \includegraphics[width=\linewidth]{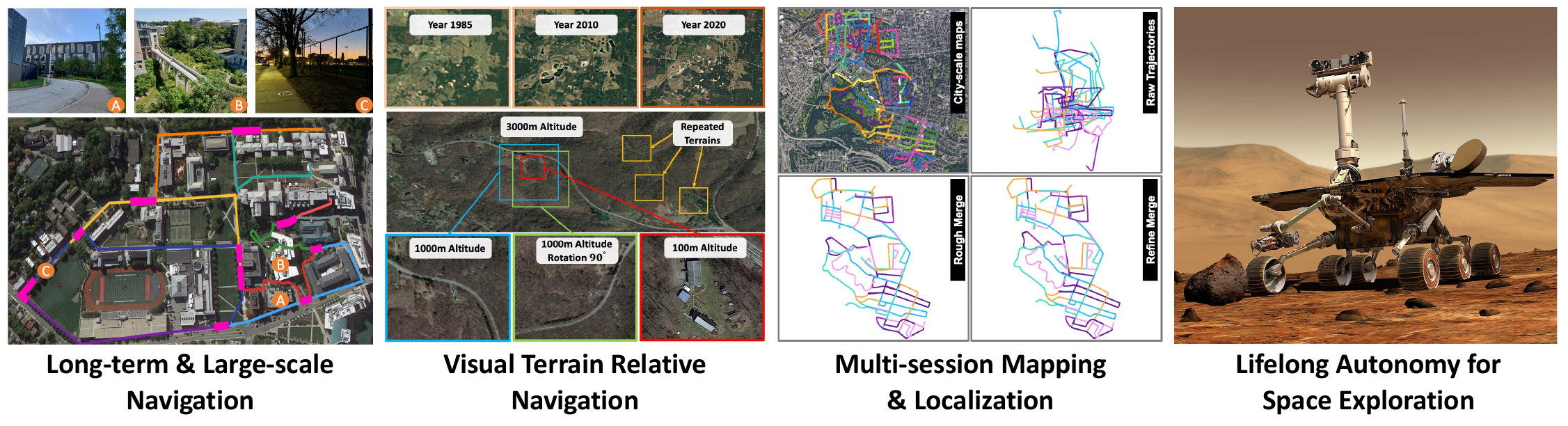}
            \end{center}
            \caption{\textbf{Application Trends for Real-world Long-term Place Recognition.}
            We list four potential application trends for reliable long-term place recognition: 1) long-term \& large-scale navigation for mobile robots, 2) visual terrain relative navigation for UAV robots, 3) multi-session mapping \& localization for single-/multi- agent systems, and 4) lifelong autonomy for space exploration.}
        \label{fig:application}
    \end{figure*}

In recent years, new kinds of loss metrics have also played an essential role in boosting place recognition performance. 
Triplet loss is a standard metric in the existing place recognition methods~\cite{NetVLAD,uy2018pointnetvlad}, which constructs the query-positive-negative pairs (based on the Euclidean distance). It stretches from the query to the positive references are minimized, and the negative inputs are maximized, as shown in Figure.~\ref{fig:loss_metric}.
Based on the triplet loss, Yin~\textit{et al.}~\cite{Yin2021spherevlad} provides a rotation triplet loss during training to improve the orientation-invariant performance.
The angular loss is another typical selection for the learning place descriptor except for triplet loss.
Barros~\textit{et al.}~\cite{loss:barros2021attdlnet} use the cosine similarity to construct the angular loss, and angular (cosine) distance is a similarity-transform-invariant metric, which is more robust than the euclidean distance in triplet loss.
However, the above loss metrics mainly consider the distance between individual cluster members with inner or external members; they do not necessarily look at the various aspects of the place recognition task explicitly. 
As shown in Figure.~\ref{fig:loss_metric}, divergence loss can explore this perspective through different ensemble modules.
And based on the above property, Yin~\textit{et al.}~\cite{yin2021pse} developed a divergence loss metric, which can enforce the place feature learning procedure care different semantic structures (i.e., trees, buildings, roads, etc.) separately.
Except for the above loss metrics, Li~\textit{et al.}~\cite{li2022rinet} treats place recognition as a classification problem, Chen~\textit{et al.}~\cite{chen2021overlapnet} utilizes a scaled sigmoid loss to estimate the overlaps between observations and achieves competitive performances.

Finally, even considering all the above approaches, developing a robust place recognition method can handle the unbounded combinations of environmental conditions and unlimited scenarios in the real-world place recognition task. 
One potential solution is lifelong learning based on place recognition.
Lifelong learning, also known as continual learning, aims at providing incrementally updated knowledge in ever-changing environments.
In~\cite{doan2020hm}, Doan~\textit{et al.} provides a Hidden Markov Model approach with a two-tiered memory management system, which can maintain the lifelong learning property by keeping necessary candidate images within the memory zones.
Kurz~\textit{et al.}~\cite{kurz2021geometry} provide a graph pruning-based method for lifelong LiDAR SLAM, which can remove  vertices and edges to keep the graph size reasonable when repetitive visit previous areas.

\subsection{Efficiency \& Robustness}
\label{sec:efficient_robust}
In real-world applications, efficiency and robustness are essential properties for long-term robotic localization.
The most straightforward place match technique uses a single-frame-based re-localization approach~\cite{mur2015orb, qin2018vins}, which can provide efficient localization for static and short-term navigation tasks.
However, real-world applications are full of dynamic changes and long-term appearance differences, and repetitive appearance patterns.
The above factors restrict the single frame-based place recognition approaches.
Instead of a single frame, sequences matching~\cite{Milford2012SeqSLAM} can improve the tolerance to local scene changes and reduce false positives with self-similarities.
SeqSLAM \cite{Milford2012SeqSLAM} and its related works \cite{SMART,Seq2}, use a brute force method to perform sequence matching.
And based on the success of SeqSLAM, new learning-based methods~\cite{Liu2019SeqLPD,Yin2019MRS_VPR,Yin2020SeqSphereVLAD,Chancan2020HybridSeqSLAM} have further boosted the place recognition performance in long-term localization tasks.
It improves the matching accuracy but is time-consuming in practice and cannot be directly used in real-time.
Thus, \cite{Fast_Yang,Fast_Siam,Fast_Peter} proposed different methods to increase the efficiency of SeqSLAM by using particle filter, approximate world's nearest neighbor, and the Hidden Markov Model. 
Methods efficiency may decrease when the number of reference sequences is enormous. 
\cite{DOSeqSLAM} and \cite{ABLE-M} improves SeqSLAM by using dynamic query sequences and binary descriptors. 
These methods' performances highly depend on the condition of the environment, which is not robust enough to perform in changing and challenging environments. 
% Then, \cite{Yin2019MRS_VPR} and \cite{Yin2020SeqSphereVLAD} introduce a framework that can balance efficiency and accuracy by improving the SeqSLAM with a coarse-to-fine searching method.
    \section{Application \& Trends}
\label{sec:application}
For future applications, there exist lots of potential directions in place for recognition.
In this section, we pick up four primary directions: 1) long-term and large-scale navigation for UGV robots, 2) visual terrain relative navigation in UAV robots, 3) multi-session SLAM, and 4) lifelong autonomy.
We analyze each direction's current status, potential outcomes, and future research trends.

\subsection{Long-term \& Large-scale Navigation}
\label{sec:app_longterm}

    Currently, the most in demand application for place recognition is the long-term and large-scale navigation, such as autonomous driving~\cite{intro:autonomous_car}, last mile delivery~\cite{intro:last_mile}, and indoor service robots~\cite{intro:service_robot}, etc.
    In the above applications, robots are expected to navigate within our daily changing indoor/outdoor environments.
    Except for the major challenges analyzed in Section.~\ref{sec:definition_challenges}, place recognition also needs to deal with the sensitive identity information~\cite{sheng2021nyu}, the high-frequent dynamic objects, and low-frequent updated geometries.

    In~\cite{torii201524}, Torii~\textit{et al.} analyze the place recognition ability under changing environmental conditions and provides a condition-invariant large-scale VPR method and extensive comparable datasets.
    For the street localization task, Liu~\textit{et al.}~\cite{liu2019lending} provide a cross-view matching for the large-scale geo-localization and encoding of both orientation and geometric information to boost the recall rates in spatial localization, which helps the city-scale robust localization.
    With the developments of deep-feature extraction~\cite{detone2018superpoint}, Sarlin~\textit{et al.}~\cite{sarlin2019coarse} provides a large-scale hierarchical localization approach, which develops a monolithic CNN structure that encodes both local geometric features and global descriptor to achieve the coarse-to-fine 6-DoF localization within large-scale environments.
    In~\cite{yin2021i3dloc}, Yin~\textit{et al.} provide the cross-modality visual localization approach within large-scale campus areas.
    Based on pre-built 3D LiDAR maps, they provide cross-domain transfer networks to extract condition-invariant features from optical inputs and learn the geometric similarity to the relative LiDAR projections in long-term navigation.
    In~\cite{rozenberszki2020lol}, Rozenberszki~\textit{et al.} integrates the LiDAR odometry and LiDAR-based place recognition method and provides a pure LiDAR-based mapping and localization system, which significantly improves the real-time localization accuracy and mapping precision in the large-scale navigation.
    And in most recent works~\cite{komorowski2021minkloc3d,vidanapathirana2021locus,vid2022logg3d}, different kinds of novel 3D place descriptors have also been investigated to boost the long-term localization performance in large-scale navigation tasks.
    To improve the localization robustness under extreme weather conditions, Hong~\textit{et al.}~\cite{hong2022radarslam} propose the entire Radar SLAM system for city-scale environments, and the experiment results also show that the place recognition performance can also support all-weather conditions.
    
    As analyzed above, long-term and large-scale navigation is currently the central research direction for place recognition, which can directly combine with most real-world robotic applications. 
    With the development of new deep feature extraction methods and different configurations of sensor modalities, this area can provide new breaking points.

\subsection{Visual Terrain Relative Navigation}
\label{sec:vtrn}
    Visual terrain-relative navigation (VTRN) is the visual localization method that can provide UAVs localization results via image registration against the target images.
    Satellite imagery is relatively easy to obtain (on earth or other planets),  which can be applied as a ground-truth image for long-term VTRN tasks.
    With the lightweight camera device and widely accessible satellite images, the VTRN approach can provide robust long-term localization in GPS-denied environments under variant conditions. 
    Due to the condition changes (Section.~\ref{sec:appearance_changes}), viewpoint differences (Section.~\ref{sec:view_diff}), and limited generalization ability (Section.~\ref{sec:generalization}), large-scale visual terrain relative navigation is still a challenging task.

    To deal with changes in environmental conditions, Mishkin \emph{et al.}~\cite{Condition:mishkin2015place} modified BoW approach~\cite{FeatureCapturer:BoW2} with multiple descriptors and adaptive thresholds to better cope with large-scale changes in environments.
    Bhavit~\emph{et al.}~\cite{Intro:ge_uav1} introduce a visual terrain navigation method where reference images are rendered from Google Earth (GE) satellite images. However, because such overhead images from GE are captured several years before test time, visual differences between reference/testing streams reduce localization accuracy.
    To deal with this issue, Mollie~\emph{et al.}~\cite{Intro:ge_uav2} utilized an Auto-encoder network to transfer raw images into overhead images, ignoring local dynamic differences or environmental condition changes.
    However, this method cannot handle illumination, weather, and seasonal changes, reducing the generalization ability for unseen conditions.
    In their recent work~\cite{VPR:SR_Season}, Anthony~\emph{et al.} provide a seasonally invariant profound transform neural network to convert seasonal images into a stable and consistent domain for visual terrain navigation.
    This method targets high-altitude flying modes, where there exist rich unchanging geometric features that persist even in different seasons. However, in lower altitudes, this method's transferability will be negatively affected by occlusions of 3D objects and highly variable lighting conditions from other times of the day.
    Another solution to deal with disturbances from environmental conditions is to match horizontal lines extracted from a query image against those rendered from digital elevation models (DEM)~\cite{Intro:DEM}.
    In~\cite{Intro:HorizonMountain}, Baatz~\emph{et al.} demonstrate terrain localization by leveraging this method.
    Similarly, Bertil~\emph{et al.}~\cite{grelsson2020gps} introduced an accurate camera localization for unmanned surface vessels (USVs) by aligning horizontal lines with coastal structures.
    A significant drawback of horizontal line-based approaches is dependence on rich 3D geometric structures, such as mountainous or coastal areas, and relative place recognition performance will reduce in homogeneous and plain environments.
    A significant drawback of horizontal line-based approaches is dependence on rich 3D geometric structures, such as mountainous or coastal areas.
    
    In general, VTRN localization is also a new research area within place recognition, which benefits from recent new domain transfer and deep feature extraction techniques.
    With the development of drone delivery and Mars exploration, the development of VTRN can be further extended.

\subsection{Multi-session SLAM}
\label{sec:app_multi}
With the development of the autonomous system, the multi-agent system has been considered to explore extensive and complex areas. It can significantly improve the operation ability of the single agent system.
And most recently, in the rover-copter system on Mars, the copter can provide better perception prediction ability in long-term rover negotiation~\cite{sasaki2020map}.
In general, collaborative multi-agent localization can benefit efficient map sharing for efficient exploration and improving perception ability for better decision making.
However, as pointed in Section.~\ref{sec:appearance_changes} and Section.~\ref{sec:view_diff}, the appearance-/viewpoint- differences from different agents will cause data association failures for multi-agent cooperation.

Van~\textit{et al.}~\cite{van2018collaborative} provide a collaborated SLAM system with compressed visual features, and enable the collaborative mapping of multi-agents on the KITTI dataset.
In~\cite{sasaki2020map}, Sasaki~\textit{et al.} provides a rover-copter-orbiter cooperative system, which utilizes the satellite images to achieve cooperated localization between the three agents; and based on the rich texture from the copter, the orbiter can generate a better path for the rover robot.
In~\cite{Merge:LAMP}, Ebadi~\textit{et al.} provide a geometric-based multi-agent SLAM system for unstable underground environments, which develops a robust filtering mechanism via the consistent incremental measurement (ICM)~\cite{mangelson2018pairwise} to reject noise data association.
However, the performance of the point-based approach is highly dependent on the robustness of 3D geometric features.
In recent work, Kimera~\cite{kimera}, Rosinol~\textit{et al.} provide a deformation graph model to merge 3D meshes between different agents, which can ensure 3D mesh consistency when used in multi-agent distributed mapping~\cite{kimera_multi}.
In~\cite{rtabmap}, Labb{\ 'e}~\textit{et al.} provide a multi-session mapping using visual appearance-based LCD methods, through which a single robot can map separate areas in different sessions without giving relative initial transformations between different trajectories.
The above methods show the potential directions in multi-agent localization. Still, we also need to note that the success of large-scale map merging is highly reliant on place recognition ability.
However, most of the above methods focus on single-agent inner data association. Still, few show promising results for large-scale multi-agent map merging tasks, where significant perspective and appearance differences between observations may exist.
In most recent work, Yin~\textit{et al.}~\cite{yin2022automerge} provide a large-scale data-association and map merging framework, which can extract viewpoint-invariant place descriptor and reject the unreliable loop-closures in the global map merging.

Multi-agent localization can be applied as the basis to support the cooperation between single-/multi- type robots, which can provide new cross-field research points in the future.

\subsection{Lifelong Autonomy}
\label{sec:lifelong_autonomy}
Recent years witnessed remarkable developments in NASA's new robotics rover (Perseverance) operation on Mars~\cite{witze2020nasa} and CNSA's teleoperated rover (Yutu-2) on the Moon~\cite{ding20222}.
But due to the costly remote operation and impractical communication, long-term and real-world autonomy ability have become necessary in future robotic systems, which will incrementally encounter unknown environments in the large-scale exploration task.
As stated in Section.~\ref{sec:definition_challenges}, robust place recognition needs to handle all the conditional and viewpoint challenges. 
The relative generalization ability is still limited when encountering new scenarios, as analyzed in Section.~\ref{sec:generalization}.
As the most fundamental module in Space/underground exploration, the localization system can directly influence the planning and decision-making strategy.
In this case, life-long place recognition has been put into an essential role in achieving long-term autonomy.
Precisely, the robots may re-visited the same place under different environmental conditions and dynamic changes.
In the above case, merging the maps and detecting environmental changes can also affect lifelong autonomy performance.
On the other hand, the robotic systems need to incrementally build up the localization knowledge from a sequentially updated data stream~\cite{lifelong_survey,CLRobot} without access to the complete datasets during the supervised/semi-supervised training manner.

Tipaldi~\textit{et al.}~\cite{tipaldi2013lifelong} provide a traditional probability-based lifelong localization approach, which combines a particle filter with a hidden Markov model to evaluate the dynamic changes of local maps.
In~\cite{zhao2021general}, Zhao~\textit{et al.} introduce a lifelong LiDAR SLAM framework for the long-term indoor navigation task, which mainly utilizes a multiple-session map to build and refine maps and parallelly maintain the memory usage via a Chow-Liu spanning tree-based optimization approach.
In real-world navigation, most SLAM methods suffer more from lower dynamic objects (i.e., parking cars) than high dynamics (i.e., moving cars).
Inspired by this observation, Zhu~\textit{et al.}~\cite{zhu2021lifelongsemi} provides a semantic mapping enhanced lifelong localization framework, which can also combine existing object detection methods to update existing maps.
However, the above methods are restricted to local scale tasks.
To tackle the large-scale applications, Yin's work~\cite{yin2022automerge} provides a long-term map merging system for city-scale environments, which utilizes an adaptive loop closure selection filter that can detect stable matches for large-scale and long-term associations.

Lifelong place feature learning is primarily evaluated against catastrophic forgetting, while most exciting place recognition methods mainly focus on short-term localization or fixed pattern localization~\cite{VPR_Bench}, catastrophic forgetting happens inevitably.
In~\cite{mactavish2017visual}, Mactavish~\textit{et al.} provides a vision-in-the-loop navigation system, which contains a visual teach-and-repeat approach to enable long-term online visual place feature learning and a multi-experience localization mechanism to help observations locate the relevant experience references.
As an extension of~\cite{mactavish2017visual}, Dall~\textit{et al.}~\cite{dall2021fast} provides a conditional teach-and-repeat visual navigation approach, which utilizes the odometry pattern as outside signals to guide the visual place feature learning.
However, in real-world localization tasks, the data stream is infinite, combining different areas under variant environmental conditions.
To tackle the scalability issue, Doan~\textit{et al.}~\cite{doan2020hm} provide a Hidden Markov Model approach with a two-tiered memory management system, which maintains an active memory and passive storage and transfers necessary candidate images between the above two memory zones to keep the lifelong property.
Doan's approach provides a new perspective on handling long-term navigation scalability issues and allows constant time/space inferencing.

Compared to other research directions, lifelong place recognition is still very young, and we can notice that there still exist many potential points in map factor and memory management for the requirements of long-term navigation tasks.
    \section{Datasets \& Evaluation}
\label{sec:data_eval}

As mentioned in Section.~\ref{sec:sensor}, the performance of different kinds of place recognition methods depends on the environmental conditions and sensor configurations.
The appropriate datasets and evaluation metrics are highly relevant to provide adequate evaluation for place recognition methods.
In this section, we will investigate the current publicly available datasets in place recognition, the star-diagram evaluation metrics given fair consideration on different place recognition properties, and support libraries for place recognition.

\begin{table*} [htbp]
    \caption{\textbf{Current long-term place recognition datasets.}}
    \centering
    \begin{tabular}{|c|c|c|c|c|c|c|}
    \hline
    \textbf{Method}
    & \textbf{Scenarios}
    & \makecell[c]{\textbf{Geographical} \\ \textbf{Coverage}}
    & \makecell[c]{\textbf{Sensor} \\ \textbf{Inputs}}
    & \makecell[c]{\textbf{Appearance} \\ \textbf{Diversity}}
    & \makecell[c]{\textbf{Viewpoint} \\ \textbf{Diversity}}
    & \makecell[c]{\textbf{Dynamic} \\ \textbf{Diversity}}
    \\
    \hline
    Nordland~\cite{sunderhauf2013we} & Train ride & $\sim 748$km & Pin-hole Camera & Four seasons & \NoMark & \NoMark\\
    \hline
    KITTI~\cite{DATASET:KITTI} & Urban Street & $\sim 1.7$km & LiDAR, Pin-hole camera & Day-time & \NoMark & \YesMark \\
    \hline
    NCLT~\cite{DATASET:NCTL} & Campus & $\sim 5.5$km & LiDAR, Panoramas Camera & Season & \YesMark & \NoMark \\
    \hline
    Oxford RobotCar~\cite{DATASET:Oxford} & Urban + Suburban & $\sim 10$km & LiDAR, Pin-hole& All kinds & \NoMark & \YesMark \\
    \hline
    Mapillary~\cite{warburg2020mapillary} & Urban + Suburban & $\sim 4,228$km & Pinhole Camera & All kinds & \YesMark & \YesMark \\
    \hline
    KITTI360~\cite{DATASET:KITTI360} & Urban Street & $\sim 73.7$km & LiDAR, Pin-hole/Panoramas Cameras & Day-time & \NoMark & \YesMark \\
    \hline
    IndoorVPR~\cite{Liu2022indoordataset} & Indoor & $15,800m^2$ & LiDAR, Pin-hole/Panoramas Cameras & Day-time & \YesMark & \NoMark\\
    \hline
    ALTO~\cite{ivan2022alto}(States) & Urban+Rural+Nature & $\sim 50$km & Top-down Pin-hole Camera & Day-time & \YesMark & \NoMark\\
    \hline
    ALITA~\cite{yin2022alita}(City)  & Urban + Terrain & $\sim 120$km & LiDAR & Day time & \YesMark & \YesMark\\
    \hline
    ALITA~\cite{yin2022alita}(Campus) & Campus & $\sim 60$km & LiDAR, Panoramas Camera & Day/Night & \YesMark & \YesMark\\ 
    \hline
    \end{tabular}
    \label{tab:public_datasets}
\end{table*}

\subsection{Public Datasets}
\label{sec:datasets}

Robust and accurate place recognition required extensive training and evaluation of datasets.
Recall the analysis in Section.~\ref{sec:sensor}, the suitable datasets for different place recognition approaches also vary based on the sensor modalities.
Though general investigations already exist on place recognition datasets~\cite{sunderhauf2013we,kang2019test,park2019radar,warburg2020mapillary,Liu2022indoordataset}, we notice that the connections from existing datasets to the sensor modalities and major challenges are still missing.
In Table.~\ref{tab:public_datasets}, we list all the relative publish available datasets with their properties.

For the most common visual place recognition, the relative datasets are mainly focus on environmental conditions(i.e., illuminations~\cite{torii201524}, weathers~\cite{yin2019multi}, seasons~\cite{sunderhauf2013we}, and dynamic objects~\cite{yamauchi1997place}) and viewpoint differences~\cite{Survey:VPR_new}.
The well-known Nordland~\cite{sunderhauf2013we} dataset provides a cross-season visual dataset with a $728km$ journey with a camera mounted on the train; this dataset has been widely applied to evaluate the condition-invariant~\cite{yin2019multi} or large-scale~\cite{Milford2012SeqSLAM} visual place recognition.
However, these datasets can not provide viewpoint-invariant analysis due to the fixed perspective, which makes the evaluation constrained under seasonal changes.
In~\cite{smith2009new}, Smith~\textit{et al.} combine the visual inputs and Laser data for campus area data collection, but both the space- and time- scales are restricted for long-term and large-scale evaluation.
Warburg~\textit{et al.}~\cite{warburg2020mapillary} provide the most long-term and large-scale visual place recognition datasets collected in $7$ years, under both urban and suburban areas, with diverse condition differences, which is highly suitable for lifelong outdoor visual place recognition.
For LiDAR place recognition, except for the most well-known KITTI~\cite{geiger2012we} datasets, the most recent autonomous driving datasets~\cite{chang2019argoverse,sun2020scalability} also give large datasets under all kinds of open-street environments.
Ramezani~\textit{et al.}~\cite{ramezani2020newer} provide campus-scale LiDAR and stereo-inertial datasets, which can be applied for VPR, LPR, and cross-modality place recognition.
For Radar-based place recognition, Sheeny~\textit{et al.}~\cite{sheeny2021radiate} provide extensive Radar datasets, which contains a variety of weather and driving scenarios to help investigate the performance of Radar device against extreme environmental conditions.
In~\cite{kim2020mulran}, Kim~\textit{et al.} provide a multimodal dataset with parallel Radar and LiDAR sensor inputs; specifically, they also offer both raw-level and image-format radar data to analyze the place recognition performance of different configurations.

Except for the above commonplace recognition tasks, other essential scenarios exist in place recognition.
The identity information is also one necessary factor in place recognition; Sheng~\textit {et al.}~\cite{sheng2021nyu} provides a new VPR dataset to analyze effects when removing all identity information within visual inputs.
For long-term visual terrain relative navigation (VTRN), Ivan~\textit{et al.}~\cite{ivan2022alto} provides a large-scale VPR dataset for UAV long-term navigation.
Specifically, they provide raw top-down visual inputs from the helicopter and the corresponding satellite images to analyze how simulation data can benefit real-world navigation.
For the purpose of multi-agent map merging, Yin~\textit{et al.}~\cite{yin2022alita} provide a large-scale LiDAR datasets, which contains $50$ trajectories and $120$ overlaps in city-scale environments, and $80$ trajectories and $150$ overlaps in campus-scale environments.
Most recently, Liu~\textit{et al.}~\cite{Liu2022indoordataset} provided intensive indoor localization datasets with RGB panorama images, LiDAR scans, and query images from smartphones for cross-modal place recognition.
Shi~\textit{et al.}~\cite{shi2020we} provide long-term indoor visual datasets with daily environmental changing and dynamics for lifelong indoor localization.

\begin{figure}[ht]
    \centering
    \includegraphics[width=0.8\linewidth]{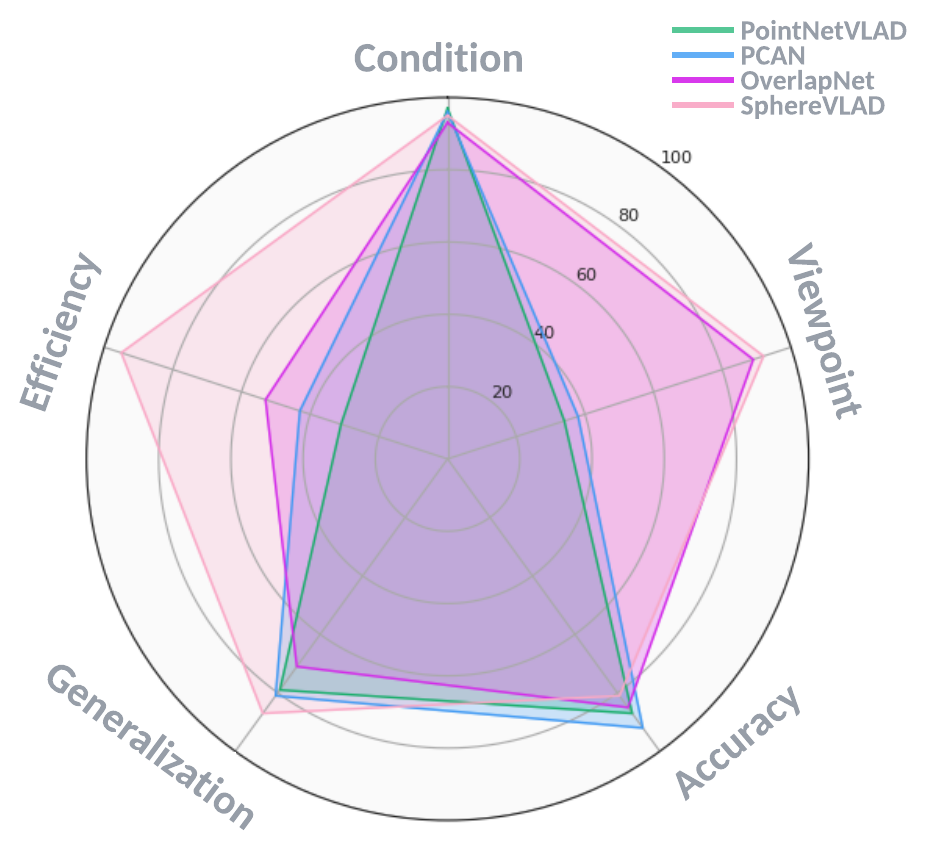}
    \caption{\textbf{Star-diagram for Place Recognition Evaluation.}
    For a comprehensive analysis, place recognition methods are analyzed under the five properties: condition-invariance, viewpoint-invariance, recognition accuracy, generalization ability, and inference efficiency.
    }
    \label{fig:star_analysis}
\end{figure}

\subsection{Star-diagram Evaluation}
\label{sec:metrics}
Recall the significant challenges in Section.~\ref{sec:definition_challenges}, the properties of place recognition are based on different factors, i.e., condition-invariant property, viewpoint-invariant property, recognition accuracy, inference efficiency, and generalization ability.
Since place recognition in the real world is a complete task, and evaluate property only on accuracy can not fully indicate the actual performance.
In previous relative benchmark~\cite{VPR_Bench}, the performance of PR methods is mainly evaluated on specific datasets, and there misses a fair quality and qualitative analysis on particular properties.
As shown in Fig.~\ref{fig:star_analysis}, we provide a star-diagram evaluation template to investigate the performance of the place recognition on,
\begin{itemize}
    \item \textit{Condition Invariant Property}: under fixed viewpoints and given sensor inputs, we analyze the place recognition accuracy (Recall rates at Top-N retrievals) under different environmental conditions (illuminations, weather, etc.)
    \item \textit{Viewpoint Invariant Property}: under fixed environmental conditions and given sensor inputs, we evaluate the distributions of place recall rates Top-N under different combinations of local translation/orientation differences.
    \item \textit{Recognition Accuracy}: under fixed environmental conditions and different local viewpoints, we analyze the final localization error (Euclidean distances) distributions of various environments near the training scenarios.
    \item \textit{Generalization ability}: under fixed environmental conditions and viewpoints, we analyze the place recognition recall rates at top-N retrievals on the unknown areas after training the network model on a fixed given dataset.
    \item \textit{Inference Efficiency}: time is one of the essential factors in real-work place recognition applications; we also explore the feature extraction time and place recognition time.
\end{itemize}

With the above star-diagram style analysis, we can provide a deep investigation of current existing PR methods or different configurations of sensor modalities.
We provide the evaluation template (Fig.~\ref{fig:star_analysis}) in
\href{https://github.com/MetaSLAM/GPRS}{GPRS} to help researchers develop new recognition methods with comprehensive perspectives.

\subsection{Supported Libraries}
\label{sec:library}
In the recent years, there also emerge lots of well maintained libraries~\cite{Seq2,humenberger2020robust,zaffar2021vpr,kim2021ltmapper,humenberger2022investigating,ivan2022alto,yin2022alita} to help develop long-term place recognition systems.
\href{https://github.com/qcr/openseqslam2}{OpenSeqSLAM2.0}~\cite{Seq2} gives a comprehensively characterizing analysis of the critical components of SeqSLAM, which helps to investigate how to maximize the performance of sequence matching.
\href{https://github.com/MubarizZaffar/VPR-Bench}{VPR-Bench}~\cite{zaffar2021vpr} provide 12 fully-integrated datasets, 10 VPR techniques, and a set of VPR evaluation metrics to help make the comparison for new methods.
\href{https://github.com/naver/kapture}{Kapture}~\cite{humenberger2020robust} is an open toolbox with an extendable data format for visual localization and structure-from-motion evaluation, which enable the VPR developments with variant requirements.
And as an extension of Kapture, \href{https://github.com/naver/kapture-localization}{Kapture-localization}~\cite{humenberger2022investigating} provides a basic localization template and can combine with different place recognition algorithms.
\href{https://github.com/gisbi-kim/lt-mapper}{LT-mapper}~\cite{kim2021ltmapper} provide a uniform lifelong mapping framework for LiDAR SLAM, which includes multi-session mapping and long-term map managements.
The LT-mapper's modular design allows it to be easily combined with new custom place descriptors.
\href{https://github.com/MetaSLAM/ALITA}{ALITA}~\cite{yin2022alita} provides the tool box for city- and campus- scale datasets, which can evaluate: 1) large-scale place recognition, 2) multi-session map merging, 3) cross-domain place recognition, and 4) long-term lifelong localization. 
And \href{https://github.com/MetaSLAM/ALTO}{ALTO}~\cite{ivan2022alto} provides the open toolbox for accessing google satellite images for raw visual inputs from UAV to evaluate large-scale visual terrain relative navigation.

    \section{Conclusion}
\label{sec:conclusion}
As stated in~\cite{yang2018grand}, simultaneous localization and mapping (SLAM) plays an essential role in mobile robotic development.
However, while the well-known visual place recognition survey paper~\cite{SURVEY:VPR} was published $6$ years ago, visual-, LiDAR-, and Radar-based place recognition and re-localization methods are still unsuccessfully applied in long-term navigation systems under varying environments.
This survey paper aims to provide a detailed review on the most recent long-term place recognition methods.
We revisit the definition of `Place Recognition' and highlight the current promising sensor modalities, major solutions for existing challenges, potential applications, trends based on reliable place recognition techniques, and publicly available datasets and libraries for long-term localization development.

With the recent development of new sensor modalities, place feature extraction from new types of sensors (high-resolution LiDAR, 360-degree camera, Radar devices) inputs have significantly improved the re-localization robustness under viewpoint and environmental conditions differences.
Most recent combinations of multi-modal sensor inputs also provide a new direction for robust long-term place recognition approaches, which can compensate for the shortcomings of using a single sensor modality. These tradeoffs can be summarized as: 1) Visual cameras capture rich semantics and textures but are not robust to appearance changes; 2) LiDAR devices provide stable 3D geometry measurements but these are low-resolution and textureless; 3) Radar devices provide robust measurements in extreme weather and occlusions but observations are often noisy.
The aforementioned different sensor configurations in Section.~\ref{sec:sensor}, we hope, will spur new research directions in long-term real-world robotics autonomy, and will also provide a new perspective on system design for new industrial products.

Deep learning-based place feature extraction has shown great success in both single- and multi-modal systems.
New types of convolution layers can be used with 360-degree sensor inputs for viewpoint-invariant place feature extraction.
New domain-transfer methods also enable sim-to-real place recognition, where robots can achieve hundreds of kilometers of re-localization with only a few kilometers of training data trajectories.
And a new type of data association mechanism, i.e., transformer networks and loss metrics, also enhances feature aggregation ability for more accurate re-localization.
And, most importantly, the increased computing power and efficiency of embedded GPUs also make online feature extraction and place re-localization achievable in real-time and in low-power consumption platforms (i.e., low-cost UGVs and UAVs).
Based on the analysis of current solutions for long-term place recognition in Section.~\ref{sec:solution}, there is still lots of space and lots of opportunities for research into generalizable place recognition approaches.

So, what can a place recognition system bring to real-world robotic research?
As analyzed in Section.~\ref{sec:application}, reliable place recognition approaches can enable long-term navigation for mobile robots in city- and campus-scale environments under challenging conditions, in complicated indoor environments with illumination variations, and even in extreme underground environments with unstructured 3D geometries; 
it can also assist with positioning systems for UAVs with the aid of publicly available satellite images, especially for space exploration (i.e., Mars, Moon), where a GPS is unavailable.
Additionally, reliable place recognition can provide stable data association across different sensor modalities or long-term time domains.
This property can provide low-cost Radar-localization on accurate 3D LiDAR maps and wearable visual localization systems for Augmented reality and virtual reality applications.
Further, with the recent developments in multi-agent systems, place recognition can also boost multi-agent cooperation in exploration and scene understanding.
Finally, given that the data-association ability of place recognition can build on place connections in both the spatial and temporal domains, it can be used to help give robots lifelong navigation and continual perception abilities for real-world applications.

In Section.~\ref{sec:data_eval}, we investigate the current publicly available datasets for long-term place recognition and provide our datasets and an evaluation benchmark to promote further research in this area.
As a community, we can further analyze new place recognition methods under different localization constraints to investigate whether or not they are suitable for real-world challenging navigation tasks.
This paper takes a sizable step toward long-term general autonomy and provides guidance in that direction.

\section{Acknowledgment}
This research was supported by grants from NVIDIA and utilized NVIDIA SDKs (CUDA Toolkit, TensorRT, and Omniverse).
This research was also supported by ARL grant NO.W911QX20D0008 and Mitsubishi Heavy Industries under grant NO.00003865.
    \bibliographystyle{IEEEtran}
    \bibliography{bible}
    %%%%%%%%%%%%%%%%%%%%%%%%%%%%%%%%%  Biography  %%%%%%%%%%%%%%%%%%%%%%%%%%%%%%%%%
%%%%%%%%%%%%%%%%%%%%%%%%%%%%%%%%%  Biography  %%%%%%%%%%%%%%%%%%%%%%%%%%%%%%%%%
%%%%%%%%%%%%%%%%%%%%%%%%%%%%%%%%%  Biography  %%%%%%%%%%%%%%%%%%%%%%%%%%%%%%%%%
\vspace{-1cm}
\begin{IEEEbiography}[{\includegraphics[width=1in,height=1.25in,clip,keepaspectratio]{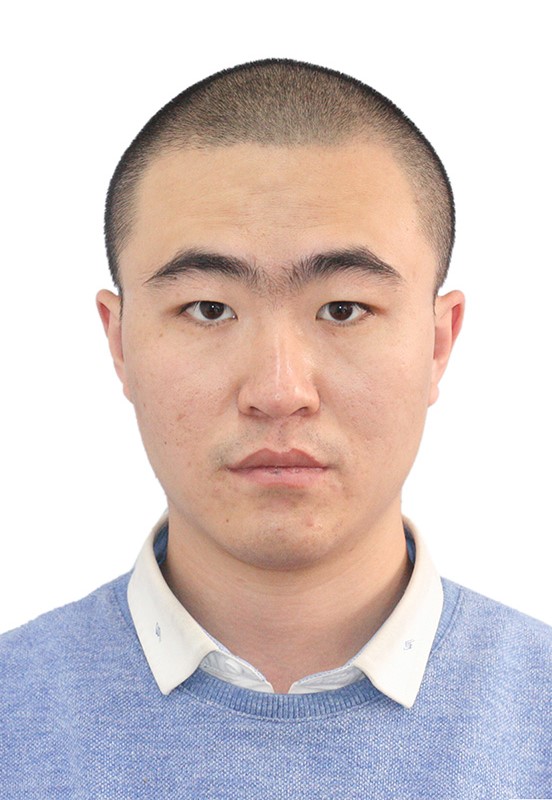}}]
    {Peng Yin} received his Bachelor degree from Harbin Institute of Technology, Harbin, China, in 2013, and his Ph.D. degree from the University of Chinese Academy of Sciences, Beijing, in 2019.
    He is a research Post-doctoral with the Department of the Robotics Institute, Carnegie Mellon University, Pittsburgh, USA.
    His research interests include LiDAR SLAM, Place Recognition, 3D Perception, and Reinforcement Learning. Dr. Yin has served as a Reviewer for several IEEE Conferences ICRA, IROS, ACC, RSS.
\end{IEEEbiography}

\begin{IEEEbiography}[{\includegraphics[width=1in,height=1.25in,clip,keepaspectratio]{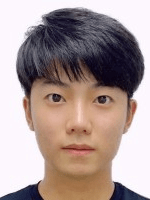}}]
    {Shiqi Zhao} received his Bachelor's degree from Dalian University of Technology, Dalian, China, in 2018, and his Master's degree from the University of California San Diego, U.S., in 2020.
    He is currently working as an intern at the Robotics Institute at Carnegie Mellon University.
    His research interests include Place Recognition, 3D Perception, and Deep Learning.
\end{IEEEbiography}

\begin{IEEEbiography}[{\includegraphics[width=1in,height=1.25in,clip,keepaspectratio]{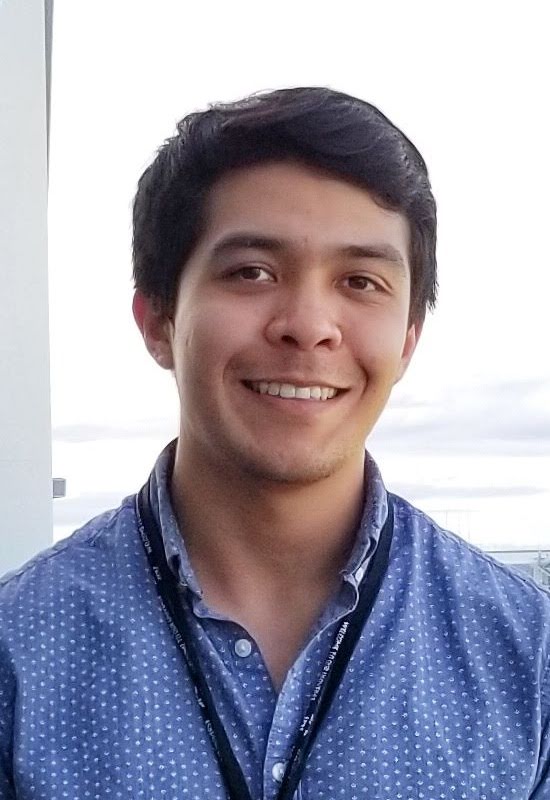}}]
    {Ivan Cisneros} received his B.S. in Electrical Engineering with a minor in Computer Science from Harvard University in 2016.
    He worked full time at NASA-JPL on several flight projects for 3 years before starting his graduate studies at Carnegie Mellon University. He is currently working on a Master's degree in Robotics within the Robotics Institute at CMU.
    His research interests include SLAM, visual localization, 3D Perception, and Deep Learning.
\end{IEEEbiography}

\begin{IEEEbiography}[{\includegraphics[width=1in,height=1.25in,clip,keepaspectratio]{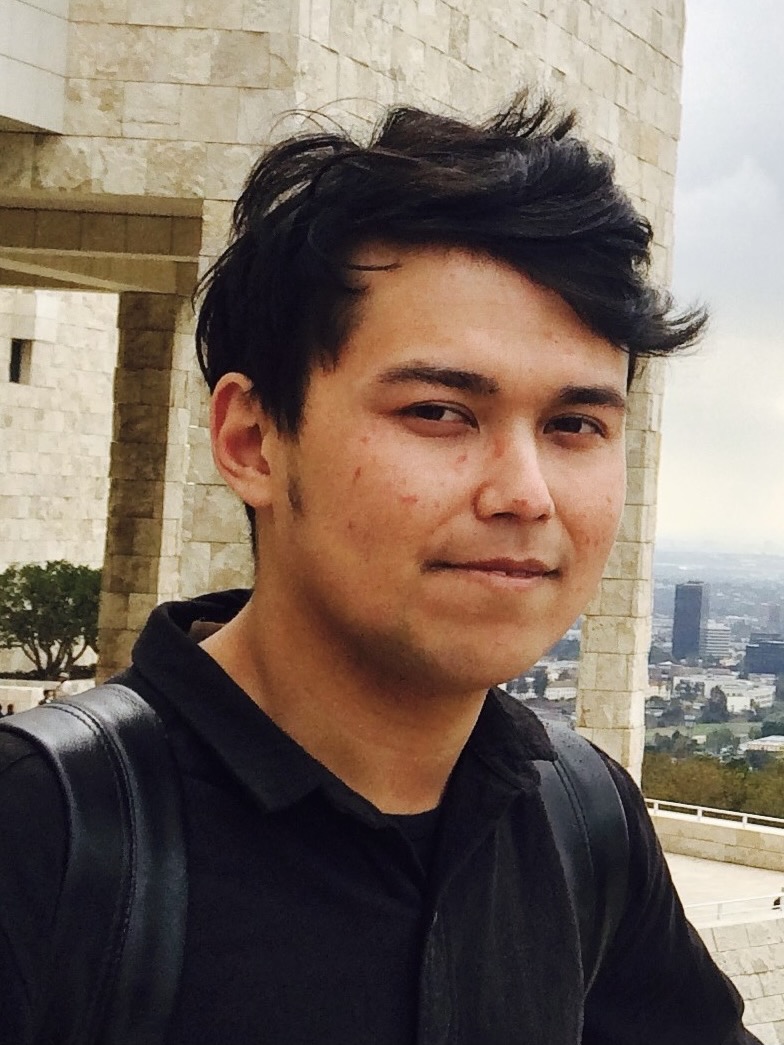}}]
    {Abulikemu Abuduweili} received his B.S. in Electrical Engineering from Peking University, China, in 2017, and his M.S. in Electrical Engineering from Peking University, China, in 2020.  
    He is currently a PhD student at Electrical Engineering and Robotics at Carnegie Mellon University, USA.
    His research interests include Deep Learning, Robotics, and Computer Vision.
\end{IEEEbiography}

\begin{IEEEbiography}[{\includegraphics[width=1in,height=1.25in,clip,keepaspectratio]{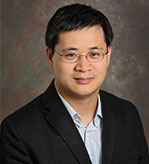}}]
    {Guoquan (Paul) Huang} (Senior Member, IEEE) received the B.S. degree in automation (electrical engineering) from the University of Science and Technology Beijing, China, in 2002, and the M.Sc. and Ph.D. degrees in computer science from the University of Minnesota--Twin Cities, Minneapolis, MN, USA, in 2009 and 2012, respectively. He currently is an Associate Professor of Mechanical Engineering (ME), Computer and Information Sciences (CIS), and Electrical and Computer Engineering (ECE), at the University of Delaware (UD), where he is leading the Robot Perception and Navigation Group (RPNG).  From 2012 to 2014, he was a Postdoctoral Associate with the MIT Computer Science and Artificial Intelligence Laboratory (CSAIL), Cambridge, MA, USA. His research interests focus on state estimation and spatial AI for autonomous vehicles and mobile devices, including probabilistic sensing, localization, mapping, perception and navigation of autonomous robots. Dr. Huang has received some honors and awards over the years, including the 2015 UD Research Award (UDRF), 2015 NASA DE Space Research Seed Award, 2016 NSF CRII Award, 2018 SATEC Robotics Delegation (one of ten US experts invited by ASME), 2018 Google Daydream Faculty Research Award, 2019 Google AR/VR Faculty Research Award, and the Winner for the IROS 2019 FPV Drone Racing VIO Competition. He was the recipient of the ICRA 2022 Outstanding Navigation Paper Award, and the Finalists of the RSS 2009 Best Paper Award and ICRA 2021 Best Paper Award in Robot Vision.

\end{IEEEbiography}

\begin{IEEEbiography}[{\includegraphics[width=1in,height=1.25in,clip,keepaspectratio]{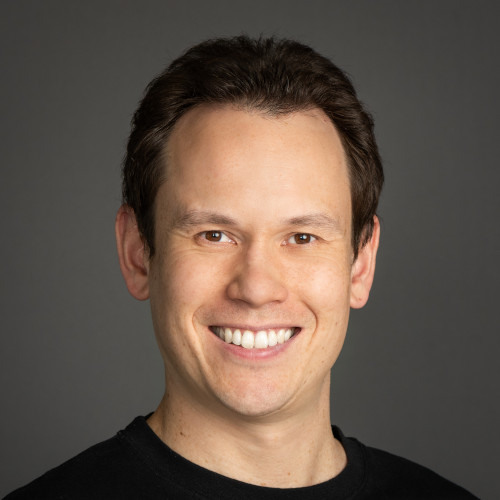}}]
    {Michael J. Milford}  (Senior Member, IEEE) received
the Bachelor of Mechanical and Space Engineering
degree and the Ph.D. degree in electrical engineering
from The University of Queensland, Brisbane, QLD,
Australia.
He was a Research Fellow on the Thinking Systems
Project with Queensland Brain Institute, in 2010.
Since 2010, he has been a Lecturer at the Queensland University of Technology (QUT), Brisbane,
QLD, Australia, where he is currently an Associate
Professor and an Australian Research Council Future
Fellow and the Chief Investigator of the Australian Centre of Excellence for
Robotic Vision. His research interests include navigation across the fields of
robotics, neuroscience, and computer vision.
Dr. Milford was a recipient of an inaugural Australian Research Council
Discovery Early Career Research Award in 2012. He has been a Microsoft
Research Faculty Fellow since 2013.
\end{IEEEbiography}

\begin{IEEEbiography}[{\includegraphics[width=1in,height=1.25in,clip,keepaspectratio]{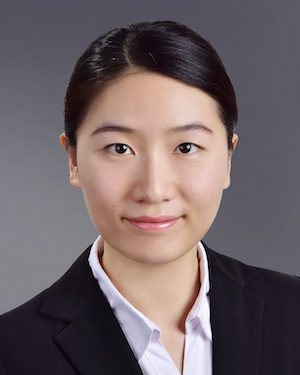}}]
    {Changliu Liu} received the B.S. degree in mechanical engineering and the B.S. degree in economics from Tsinghua University, China, in 2012, the M.S. degree in mechanical engineering, and the M.A. degree in mathematics from University of California, Berkeley, U.S.A., in 2014 and 2016 respectively, and the Ph.D. degree in mechanical engineering from University of California, Berkeley, U.S.A., in 2017. She is an assistant professor at the Robotics Institute at Carnegie Mellon University. Her research interests lie in the design and verification of intelligent systems with applications to manufacturing and transportation. She received NSF Career Award, Amazon Research Award, and Ford URP Award.
\end{IEEEbiography}

% \begin{IEEEbiography}[{\includegraphics[width=1in,height=1.25in,clip,keepaspectratio]{Biography/zhang_ji.jpg}}]
%     {Ji Zhang} received his Ph.D. in Robotics from Carnegie Mellon University in 2017.
%     Ji Zhang is a Systems Scientist at the Robotics Institute at Carnegie Mellon University, where he leads in the development of a series of autonomous navigation algorithms. His work was ranked \#1 on the odometry leaderboard of KITTI Vision Benchmark between 2014 and 2021. He founded Kaarta, Inc, a CMU spin-off commercializing 3D mapping \& modeling technologies, and stayed with the company for 4 years as chief scientist.
%     His research interests are in robotic navigation, spanning localization, mapping, planning, and exploration.
% \end{IEEEbiography}

\begin{IEEEbiography}[{\includegraphics[width=1in,height=1.25in,clip,keepaspectratio]{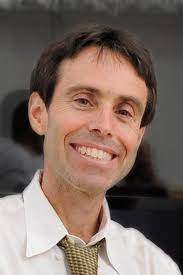}}]
    {Howie Choset} received the B.S. Eng. degree in computer science and the B.S. Econ. degree in entrepreneurial management from the University of Pennsylvania (Wharton), Philadelphia, PA, USA, both in 1990, the M.S. and Ph.D. degrees in mechanical engineering from California Institute of Technology (Caltech), Pasadena, CA, USA, in 1991 and 1996, respectively.
    He is currently a Professor of Robotics with the Carnegie Mellon University, Pittsburgh, PA, USA. His research group reduces complicated high dimensional problems found in robotics to low-dimensional simpler ones for design, analysis, and planning. 
    % Prof. Choset was elected as one of the top 100 innovators in the world under 35, by the MIT Technology Review, in 2002. In 2014, Popular Science selected his medical robotics work as the Best of What’s New in Health Care. 
\end{IEEEbiography}

\vspace{-1cm}
\begin{IEEEbiography}[{\includegraphics[width=1in,height=1.25in,clip,keepaspectratio]{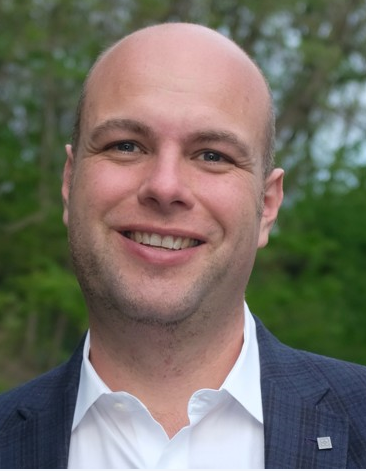}}]
    {Sebastian Scherer} received his B.S. in Computer Science, M.S. and Ph.D. in Robotics from CMU in 2004, 2007, and 2010. 
    Sebastian Scherer is an Associate Research Professor at the Robotics Institute at Carnegie Mellon University. His research focuses on enabling autonomy for unmanned rotorcraft to operate at low altitude in cluttered environments. He is a Siebel scholar and a recipient of multiple paper awards and nominations, including AIAA@Infotech 2010 and FSR 2013. His research has been covered by the national and internal press including IEEE Spectrum, the New Scientist, Wired, der Spiegel, and the WSJ. 
\end{IEEEbiography}

\endgroup

\end{document}